\documentclass[letterpaper, 10 pt, conference]{ieeeconf}  % Comment this line out if you need a4paper

\IEEEoverridecommandlockouts                              % This command is only needed if 
\usepackage[pdftex]{graphicx}
\usepackage{amsmath}
\usepackage{amssymb}
\usepackage{mathtools}

\usepackage[font=footnotesize]{caption}
\usepackage[font=footnotesize, subrefformat=parens]{subcaption}
\usepackage[ruled,linesnumbered,noend]{algorithm2e}
\SetAlCapFnt{\footnotesize}
\SetAlgorithmName{Alg.}{Alg.}{List of Algorithms}

\usepackage{cite}
\usepackage{comment}

\usepackage{booktabs}

\newcommand{\argmax}{\mathop{\rm arg~max}\limits}

\title{\LARGE \bf
Gaussian Process Self-triggered Policy Search \\ 
in Weakly Observable Environments
}

\author{Hikaru Sasaki$^{1}$, Terushi Hirabayashi$^{2}$, Kaoru Kawabata$^{2}$, and Takamitsu Matsubara$^{1}$% <-this % stops a space
\thanks{$^{1}$H. Sasaki and T. Matsubara are with the Division of Information Science, Graduate School of Science and Technology, Nara Institute of Science and Technology (NAIST), Japan. (email: {\tt\footnotesize \{sasaki.hikaru.rw3, takam-m\}@is.naist.jp)}}
\thanks{$^{2}$T. Hirabayashi and K. Kawabata are with Hitachi Zosen Corporation, Japan. (email: {\tt\footnotesize \{hirabayashi, kawabata\_k\} @hitachizosen.co.jp})}
}

\begin{document}

\maketitle
\thispagestyle{empty}
\pagestyle{empty}

%%%%%%%%%%%%%%%%%%%%%%%%%%%%%%%%%%%%%%%%%%%%%%%%%%%%%%%%%%%%%%%%%%%%%%%%%%%%%%%%
\begin{abstract}
The environments of such large industrial machines as waste cranes in waste incineration plants are often \textit{weakly observable}, where little information about the environmental state is contained in the observations due to technical difficulty or maintenance cost (e.g., no sensors for observing the state of the garbage to be handled). Based on the findings that skilled operators in such environments choose predetermined control strategies (e.g., grasping and scattering) and their durations based on sensor values, %thereby improving the robustness of their actions, 
we propose a novel non-parametric policy search algorithm: Gaussian process self-triggered policy search (GPSTPS). 
GPSTPS has two types of control policies: action and duration. A gating mechanism either maintains the action selected by the action policy for the duration specified by the duration policy or updates the action and duration by passing new observations to the policy; therefore, it is categorized as \textit{self-triggered}.
GPSTPS simultaneously learns both policies by trial and error based on sparse GP priors and variational learning to maximize the return. 
To verify the performance of our proposed method, we conducted experiments on garbage-grasping-scattering task for a waste crane with weak observations using a simulation and a robotic waste crane system.
As experimental results, the proposed method acquired suitable policies to determine the action and duration based on the garbage's characteristics.
\end{abstract}

%%%%%%%%%%%%%%%%%%%%%%%%%%%%%%%%%%%%%%%%%%%%%%%%%%%%%%%%%%%%%%%%%%%%%%%%%%%%%%%%
\section{INTRODUCTION}

Policy search reinforcement learning has received much attention as a method for learning control policies for robots \cite{chatzilygeroudis2020}.
In particular, policy search using non-parametric policies such as Gaussian process (GP) regression is effective for complex tasks due to its nonlinear and stochastic properties \cite{hoof2017, sasaki2021}.
Moreover, the complexity of its model can be automatically adjusted using data \cite{rasmussen2006}.
Therefore, compared to parametric policies such as neural networks, policy search using GP-policy reduces the effort of model design, resulting in high sample efficiency \cite{hoof2017, sasaki2021}.

However, policy search is difficult for automating heavy industrial machinery in actual workplaces. 
Their control systems still have to rely on complicated system-specific models created by humans \cite{ramli2017}. 
In particular, two major challenges must be tackled in policy search for industrial machines: 
\begin{enumerate}
\item Obtaining trial and error data by a large and heavy industrial machine is very costly. 
\item Since the environments of industrial machines are \textit{weakly observable}, little information about the state is contained in observations due to technical difficulties or maintenance costs.
\end{enumerate}

More specifically, a waste crane, remotely controlled by a skilled operator at a waste incineration plant, does not have sensors that collect information about the garbage's state, although its characteristics have diversity and changing trends that depend on the day of the week and the season (size and hardness of each element, moisture content, stickiness, etc.) 
\cite{elasri2004,sasaki2020}.
The crane is only equipped with a weight sensor to observe information for grasped garbage by the bucket.
Although the operators can roughly see the whole garbage pit from the control room, due to severe occlusion they generally cannot see the garbage around the bucket, especially when it lands the waste and executes a grasping motion. 

\begin{figure}
    \centering
    \includegraphics[width=0.9\hsize]{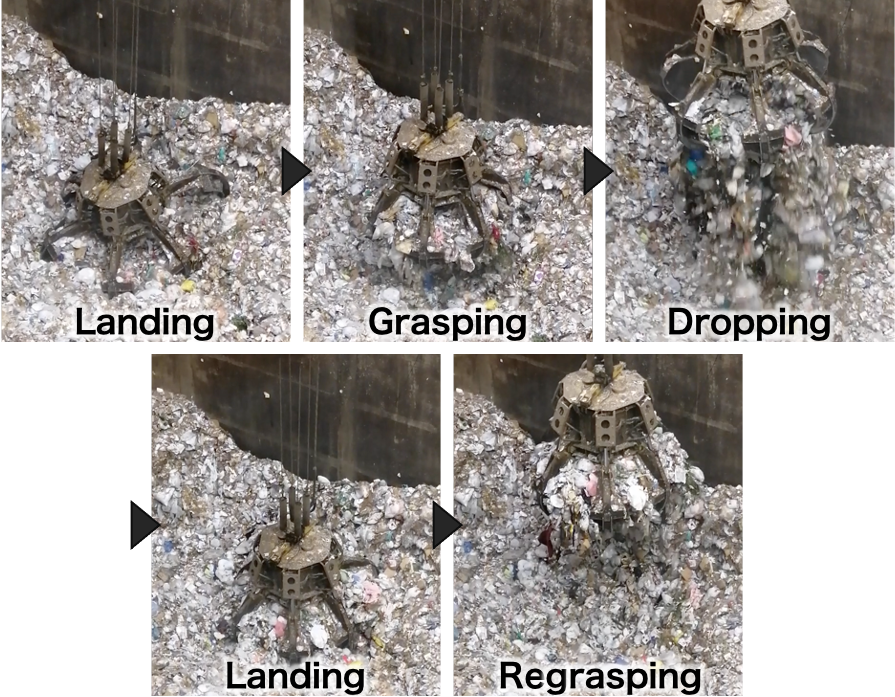} 
    \caption{Garbage grasping by an actual waste crane by a human operator}
    \label{fig_actual_waste_crane}
\end{figure}

Such a control problem can generally be viewed as a partially observed Markov decision process (POMDP).
POMDP solution methods aim to estimate beliefs (probability distributions of states) from observed data to decide actions \cite{shani2013}.
However, such methods are intractable in such environments as garbage cranes, where the observations contain very little information.
skilled operators can control such machines and achieve high performance. % with such scant information about the environmental state.
Their behavior seems different from general sensory feedback control that selects an action based on sensor values at regular time intervals (Fig. \ref{fig_actual_waste_crane}), which is a typical policy model in policy search and RL (e.g., \cite{tsurumine2019a,sasaki2021}). 
This situation can be attributed to the weakness of the observations; if the value of the weight sensor does not change significantly, the operator will not be able to select a different action.
Instead, selecting a predetermined control strategy (e.g., grasping or scattering) and a duration based on sensor values creates robust behavior and stabilizes it even with weak observations.
Such an approach can reduce the dependency of the policy on sensor values and the frequency of references.
However, these policies need to be adjusted according to the garbage's characteristics.
Such thoughts motivated us to explore a novel policy search framework with a specific policy applicable in weakly observable environments in a sample-efficient way.  

In this paper, we propose GP self-triggered policy search (GPSTPS), which is a novel non-parametric policy search algorithm.
GPSTPS has two types of control policies: action and duration. 
The gating mechanism either maintains the action selected by the action policy for the duration specified by the duration policy or updates the action and duration by passing new observations to the policy; therefore, it is described as \textit{self-triggered}.
GPSTPS simultaneously learns both policies by trial and error based on sparse GP priors \cite{titsias2009} and variational learning \cite{sasaki2021} to maximize the return. 
We experimentally verified the performance of our proposed method on a garbage-grasping-scattering task with a waste crane with weak observations using a simulation and a robotic waste crane system.
Our experimental results suggest that our method acquired suitable policies to determine the action and duration based on the garbage's characteristics.

\section{RELATED WORK}
\subsection{Crane and Bucket Automation} % for Partially Observable Environment}
Crane automation is conventionally implemented with a complicated model of a crane system \cite{ramli2017}.
Robust control over disturbances has also been explored by modeling such as wind \cite{abdullahi2018} and waves 
\cite{qian2017}.
For such difficult-to-model objects as soil, previous work designed robust movements for an excavator 
\cite{sotiropoulos2019}.
In a data-driven approach, learning methods were proposed that are predictive models for excavation performance
\cite{lu2021}
and imitation learning methods for wheel loaders \cite{dadhich2019}.
Such works designed elaborate models and learned models with much sensor information; however, such approaches may not be suitable in environments that have weak observations.
In this paper, we propose a data-driven policy search method for weakly observable environments.

\subsection{Aperiodic Action Update}
In the control theory domain, event and self-triggered controls, which are aperiodic action update methods, have been proposed to reduce communication costs between the controller and the plant \cite{heemels2012}.
Event- and self-triggered control consists of a controller and triggering mechanism. 
In event-triggered control, an action update event is triggered when a current observation violates the triggering condition.
Thus, event-triggered control is robust to disturbance, however, it assumes that the full state information is available.
% On the other hand, 
In self-triggered control, the next action update time is precomputed when an action is updated.
Self-triggered control computes the next action update time using the model.

In this paper, we employ self-triggered control-like duration determination by GP-based duration policy since the proactive duration does not need to capture small changes in observations and can determine a duration from observation.

\subsection{Reinforcement Learning with extended MDP}
Complex tasks that cannot be solved by general RL are formulated with extended MDP.
One extension of MDP is the semi-Markov decision process (SMDP).
The option framework reinforcement learning is one of the methods for SMDP, that temporally abstract task to sub-tasks related to option \cite{sutton1999, daniel2016, wulfmeier2021}.
In these learning methods, the task is segmented into sub-tasks and sub-policies are learned to achieve related sub-tasks.
Those methods execute complex tasks by switching a suitable option aperiodically.
The option switching is happened by a termination model that is similar to a triggering condition in event-triggered control.

Tasks that are difficult to make decisions based on current observations due to insufficient information are formulated as partially observable Markov decision processes (POMDP).
For POMDP, RL methods have introduced a belief that indicates a current state cannot be observed and is inferred from the past state and action series to determine action \cite{shani2013}.

In an environment with less information, observable information is inadequate for inferring additional information like belief and option.
Therefore, our proposed method assumes designing predetermined control strategies instead of inference and verifies the effectiveness of the application.

\section{POLICY SEARCH}
In this section, we describe model-free policy search as preliminary to derive our proposed method.
Policy search is an algorithm that acquires a policy that maximizes the expected return in the environment formulated by the Markov decision process.
The policy is learned by repeating the exploration and improvement as episodic reinforcement learning.
Episode $d = \{\mathbf s_1, a_1, \cdots,\mathbf s_{T}, a_{T},\mathbf s_{T+1}\}$, which is a series of state $\mathbf s_t$ and action $a_t$ is collected in a trial and error by the policy $\pi$.
A reward $r(\mathbf s_t, a_t)$ is calculated using a state and an action.
Here we assume the following derivation where the action is one-dimensional for clarity without a loss of generality.
Let the initial state probability and the state transition probability be $p(\mathbf s_1)$ and $p(\mathbf s_{t+1}\mid\mathbf s_t, a_t)$, and express the probability of episode $p(d\mid\pi)$ as:
\begin{align}
    p(d\mid\pi) = p(\mathbf s_1)\prod_{t=1}^T \pi(a_t\mid\mathbf  s_t)p(\mathbf s_{t+1}\mid\mathbf  s_t, a_t).
\end{align}
The policy is improved by calculating it to maximize the expected return $J(\pi)=\int R(d) p(d\mid\pi) ~\mathrm d d$ using return function $R(d)=\sum_{t=1}^T r(\mathbf s_t, a_t)$ and probability $p(d\mid\pi)$:
\begin{align}
    \pi^* &= \argmax_\pi J(\pi). 
    % J(\pi) &= \int R(d) p(d\mid\pi) ~\mathrm d d.
\end{align}

\section{GAUSSIAN PROCESS SELF-TRIGGERED POLICY SEARCH}
In this section, we propose a self-triggered policy search with GP as a policy model.
% An action and duration selection by the self-triggered policy in task execution is indicated in Alg. \ref{alg_self_triggered_task_execution}.
We introduce the self-triggered policy model to policy search and derive the policy update low based on variational policy search \cite{sasaki2021} by formulating the expected return as the return-weighted marginal likelihood.

\subsection{Problem Formulation}
Self-triggered policy search has an execution duration of $\tau_t$, which indicates the time to continue action $a_t$, and binary gating variable $o_t$, which indicates when the action and its duration time are updated by the policy at time $t$.
The action and duration at each time are modeled:
\begin{align}
    p(a_t \mid \mathbf s_t, o_t, a_{t-1}, \pi_a) &=
    \begin{cases}
        \pi_a(a_t\mid\mathbf s_t) &\mathrm{if}~ o_t=1 \\
        \delta(a_t-a_{t-1})       &\mathrm{else}
    \end{cases}, \\
    p(\tau_t \mid \mathbf s_t, o_t, \tau_{t-1}, \pi_\tau) &=
    \begin{cases}
        \pi_\tau(\tau_t\mid \mathbf s_t) &\mathrm{if}~ o_t=1 \\
        \delta(\tau_t-\tau_{t-1}+1)      &\mathrm{else}
    \end{cases}.
\end{align}
These two distributions include action policy $\pi_a$ and duration policy $\pi_\tau$.
Gating variable $o_t$ indicates action update or maintenance, and when $o_t=1$, the action and the duration are updated by each policy.
When $o_t=0$, the previous action is continued and the duration is decremented.
Duration time $\tau_t$ indicates the remaining execution duration of action $a_t$.
Gating variable $o_t$ is determined by duration $\tau_{t-1}$:
\begin{align}
    p(o_t\mid \tau_{t-1}) =
    \begin{cases}
        \delta(o_t-1) &\mathrm{if}~~ \tau_{t-1}=1 \\
        \delta(o_t) &\mathrm{else}
    \end{cases}.
\end{align}
The gating variable becomes $o_t=1$ when the duration is $\tau_{t-1}=1$.

In self-triggered policy search, the episode is extended as $d_s = \{\mathbf s_1, a_1, \tau_1, o_1, \cdots, \mathbf s_{T}, a_{T}, \tau_T, o_T, \mathbf s_{T+1}\}$.
The extended episode's probability is described:
\begin{align}
    &p(d_s\mid\pi_a,\pi_\tau) = p(\mathbf s_1)\prod_{t=1}^T p(a_t\mid \mathbf s_t,o_t, a_{t-1},\pi_a)\cdot \nonumber \\
    &\hspace{5mm}p(\tau_t\mid \mathbf s_t,o_t, \tau_{t-1},\pi_\tau)p(o_t\mid\tau_{t-1})p(\mathbf s_{t+1}\mid \mathbf s_t, a_t).
\end{align}
Self-triggered policy search uses the episode's probability to calculate the expected return $J(\pi_a,\pi_\tau) = \int R(d_s)p(d_s\mid\pi_a,\pi_\tau)\mathrm d d$ and learns action policy $\pi_a$ and duration policy $\pi_\tau$:
\begin{align}
    % J(\pi_a,\pi_\tau) &= \int R(d_s)p(d_s\mid\pi_a,\pi_\tau)\mathrm d d, \\
    \pi_a^*, \pi_\tau^* &= \argmax_{\pi_a,\pi_\tau} J(\pi_a,\pi_\tau).
\end{align}

\subsection{Sparse Gaussian Process Policy Models}
GPs are employed as a policy model in this paper.
The nonlinear functions of the action and duration policies are defined as $f$ and $g$, and GPs are set as their priors:
% $f \sim \mathcal{GP}(m_f, \mathrm k(\mathbf s, \mathbf s'))$,
% $g \sim \mathcal{GP}(m_g, \mathrm k(\mathbf s, \mathbf s'))$,
\begin{align}
    f \sim \mathcal{GP}(m_f, \mathrm k(\mathbf s, \mathbf s')), 
    g \sim \mathcal{GP}(m_g, \mathrm k(\mathbf s, \mathbf s')),
\end{align}
where $\mathrm k(\cdot, \cdot)$ is a kernel function and $m_f$ and $m_g$ are mean of each GP.
We assume a Gaussian distribution for the action and duration:
\begin{align}
    % p(a_t\mid f_t) &= \mathcal{N}(a_t\mid f_t, \sigma^2_f), \\
    % p(\tau_t\mid g_t) &= \mathcal{N}(\tau_t\mid g_t, \sigma^2_g),
    p(a_t | f_t) = \mathcal{N}(a_t | f_t, \sigma^2_f),  ~~
    p(\tau_t | g_t) = \mathcal{N}(\tau_t | g_t, \sigma^2_g),
\end{align}
where $f_t=f(\mathbf s_t)$, $g_t=g(\mathbf s_t)$, $\sigma^2_f$ and $\sigma^2_g$ are the variances of each Gaussian distribution.

To reduce the computational complexity of GPs used as the prior distribution of the nonlinear function, pseudo outputs $\bar{\mathbf f}$ and $\bar{\mathbf g}$ corresponding to pseudo input $\bar{\mathbf s}$ \cite{titsias2009} are introduced and the prior distributions are set:
\begin{align}
    % p(\bar{\mathbf f}\mid \bar{\mathbf s}) &= \mathcal N(\bar{\mathbf f}\mid m_f\mathbf 1, \mathbf K_{\bar{\mathbf s}}),\\
    % p(\bar{\mathbf g}\mid \bar{\mathbf s}) &= \mathcal N(\bar{\mathbf g}\mid m_g\mathbf 1, \mathbf K_{\bar{\mathbf s}}),
    p(\bar{\mathbf f} | \bar{\mathbf s}) = \mathcal N(\bar{\mathbf f} | m_f\mathbf 1, \mathbf K_{\bar{\mathbf s}}),
    p(\bar{\mathbf g} | \bar{\mathbf s}) = \mathcal N(\bar{\mathbf g} | m_g\mathbf 1, \mathbf K_{\bar{\mathbf s}}),
\end{align}
where $\mathbf K_{\bar{\mathbf s}}=\mathrm k(\bar{\mathbf s},\bar{\mathbf s})$ is the kernel gram matrix of pseudo input $\bar{\mathbf s}$.
The distributions of nonlinear function outputs $f_t$ and $g_t$ are represented by the following Gaussian distributions as the GP regression of the distribution of the pseudo outputs:
\begin{align}
    p(f_t\mid \mathbf s_t, \bar{\mathbf f}) &= \mathcal N(f_t\mid \mathbf k_{\mathbf s_t,\bar{\mathbf s}}\mathbf K_{\bar{\mathbf s}}^{-1}(\bar{\mathbf f}-m_f)+m_f, \lambda_t), \\
    p(g_t\mid \mathbf s_t, \bar{\mathbf g}) &= \mathcal N(g_t\mid \mathbf k_{\mathbf s_t,\bar{\mathbf s}}\mathbf K_{\bar{\mathbf s}}^{-1}(\bar{\mathbf g}-m_g)+m_g, \lambda_t),
\end{align}
where $\lambda_t = k_{\mathbf s_t}-\mathbf k_{\mathbf s_t,\bar{\mathbf s}}\mathbf K_{\bar{\mathbf s}}^{-1}\mathbf k_{\mathbf s_t,\bar{\mathbf s}}^T$, $k_{\mathbf s_t}=\mathrm k(\mathbf s_t,\mathbf s_t)$, and $\mathbf k_{\mathbf s_t,\bar{\mathbf s}}=\mathrm k(\mathbf s_t,\bar{\mathbf s})$.

The probability of an episode of self-triggered policy search using GPs as a policy model is calculated as follows:
\begin{align}
    &p(d_s,\mathbf f, \mathbf g \mid \bar{\mathbf f}, \bar{\mathbf g})=p(\mathbf s_1)\prod_{t=1}^T p(a_t\mid f_t, o_t, a_{t-1})p(f_t\mid \mathbf s_t, \bar{\mathbf f})\cdot \nonumber \\
    &p(\tau_t | g_t, o_t, \tau_{t-1})p(g_t | \mathbf s_t, \bar{\mathbf g})p(o_t | \tau_{t-1})p(\mathbf s_{t+1} | \mathbf s_t,  a_t),
    \label{eq_expected_return}
\end{align}
where $\mathbf f=\{f_t\}_{t=1}^T$ and $\mathbf g=\{g_t\}_{t=1}^T$.

\subsection{Variational Learning for Policy Improvement}
We derive GPSTPS's policy improvement as a return-weighted marginal likelihood maximization problem using explored data.
%The posterior distribution of the nonlinear functions in the two policy models, kernel parameters $\theta$, and pseudo inputs $\bar{\mathbf f}$ and $\bar{\mathbf g}$ are optimized by variational learning.
The posterior distribution of the nonlinear functions $f$ and $g$, kernel parameters $\theta$, and pseudo inputs $\bar{\mathbf f}$ and $\bar{\mathbf g}$ are optimized by variational learning.

With the episode's probability using the GP policy model in (\ref{eq_expected_return}), the expected return is calculated:
\begin{align}
    J = \int R(d_s)p(d_s, \mathbf f, \mathbf g \mid \bar{\mathbf f}, \bar{\mathbf g})p(\bar{\mathbf f})p(\bar{\mathbf g})\mathrm dd_s\mathrm d\mathbf f\mathrm d\bar{\mathbf f}\mathrm d\mathbf g\mathrm d\bar{\mathbf g}.
\end{align}
This equation cannot be analytically solved due to the complexity of $d_s$, $\bar{\mathbf f}$, and $\bar{\mathbf g}$.
Therefore, by introducing the distribution of the data sample $p_\mathrm{old}(d_s)$, expected return $J_\mathrm{old}$ by that distribution, and variation distribution $q(d_s,\mathbf f,\bar{\mathbf f},\mathbf g, \bar{\mathbf g}) = p_\mathrm{old}(d_s) p(\mathbf f \mid \bar{\mathbf f}) q (\bar{\mathbf f}) p(\mathbf g \mid \bar{\mathbf g}) q(\bar{\mathbf g})$, the lower bound of expected return $\log J_L$ is derived:
\begin{align}
    &\log \frac{J(\theta)}{J_\mathrm{old}} \geq \int \frac{R(d_s)}{J_\mathrm{old}}q(d_s, \mathbf f, \bar{\mathbf f}, \mathbf g, \bar{\mathbf g})\cdot \nonumber \\
    &~~~~~~~~~~~~~~~~~\log \frac{p(d_s\mid \bar{\mathbf f}, \bar{\mathbf g})p(\bar{\mathbf f})p(\bar{\mathbf g})}{q(d_s, \mathbf f, \bar{\mathbf f}, \mathbf g, \bar{\mathbf g})}\mathrm dd_s\mathrm d\mathbf f\mathrm d\bar{\mathbf f}\mathrm d\mathbf g\mathrm d\bar{\mathbf g} \nonumber\\
    &= \int \frac{R(d_s)}{J_\mathrm{old}}p_\mathrm{old}(d_s)p(\mathbf f\mid\bar{\mathbf f})q(\bar{\mathbf f})\log \frac{p(\mathbf a\mid\mathbf f)p(\bar{\mathbf f})}{q(\bar{\mathbf f})}\mathrm dd_s\mathrm d\mathbf f\mathrm d\bar{\mathbf f} \nonumber \\
    &+ \int \frac{R(d_s)}{J_\mathrm{old}}p_\mathrm{old}(d_s)p(\mathbf g\mid\bar{\mathbf g})q(\bar{\mathbf g})\log \frac{p(\boldsymbol\tau\mid\mathbf g)p(\bar{\mathbf g})}{q(\bar{\mathbf g})}\mathrm dd_s\mathrm d\mathbf g\mathrm d\bar{\mathbf g} \nonumber \\
    &+C\nonumber \\
    % &= \int \frac{R(d_s)}{J_\mathrm{old}}q(d_s,\mathbf f,\bar{\mathbf f},\mathbf g, \bar{\mathbf g})\cdot \nonumber\\
    % &\left\{\log \frac{p(\mathbf a\mid\mathbf f)p(\bar{\mathbf f})}{q(\bar{\mathbf f})} + \log \frac{p(\boldsymbol\tau\mid\mathbf g)p(\bar{\mathbf g})}{q(\bar{\mathbf g})}\right\}\mathrm dd_s\mathrm d\mathbf f\mathrm d\bar{\mathbf f}\mathrm d\mathbf g\mathrm d\bar{\mathbf g} +C\nonumber \\
    &=\log J_L(\theta, q),
\end{align}
where $\mathbf a=\{a_t\}_{t=1}^T$ and $\boldsymbol{\tau}=\{\tau_t\}_{t=1}^T$.
Each policy model can be learned using variational policy search \cite{sasaki2021} since the lower bound is a sum of the marginal likelihood for each policy.

\section{EXPERIMENT}
We applied GPSTPS to a garbage-grasping-scattering task by simulation and a robotic waste crane system to investigate its effectiveness.
% We verified that GPSTPS appropriately learned the action and duration policies based on the garbage's characteristics.

\begin{figure}
    \centering
    \begin{minipage}[b]{0.75\linewidth}
        \centering
        \includegraphics[width=\hsize]{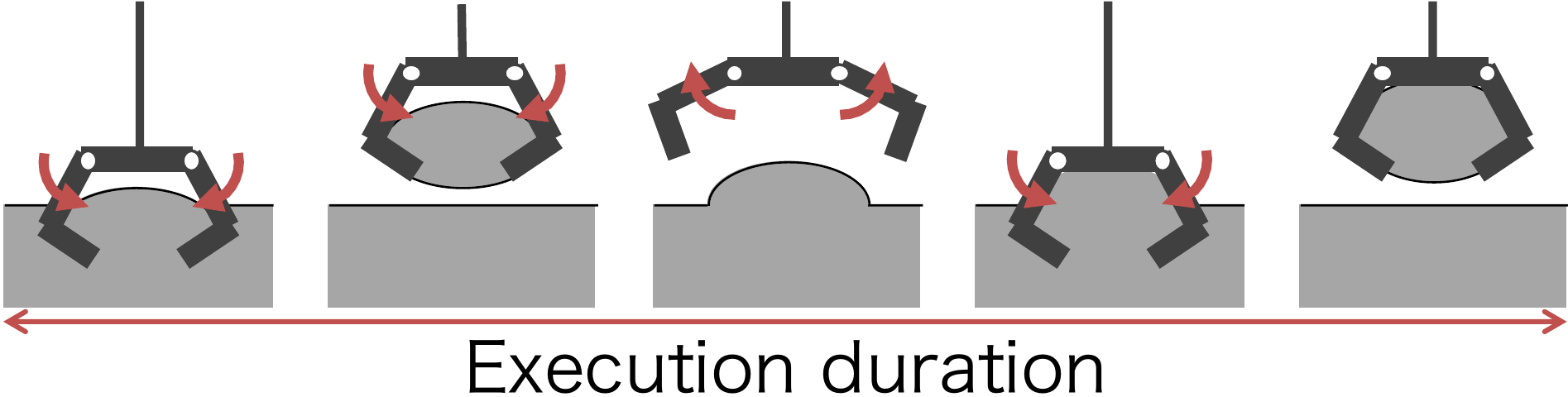} 
        \subcaption{Grasping}
        \label{fig_problem_grasp}
    \end{minipage}\\
    \begin{minipage}[b]{0.75\linewidth}
        \centering
        \includegraphics[width=\hsize]{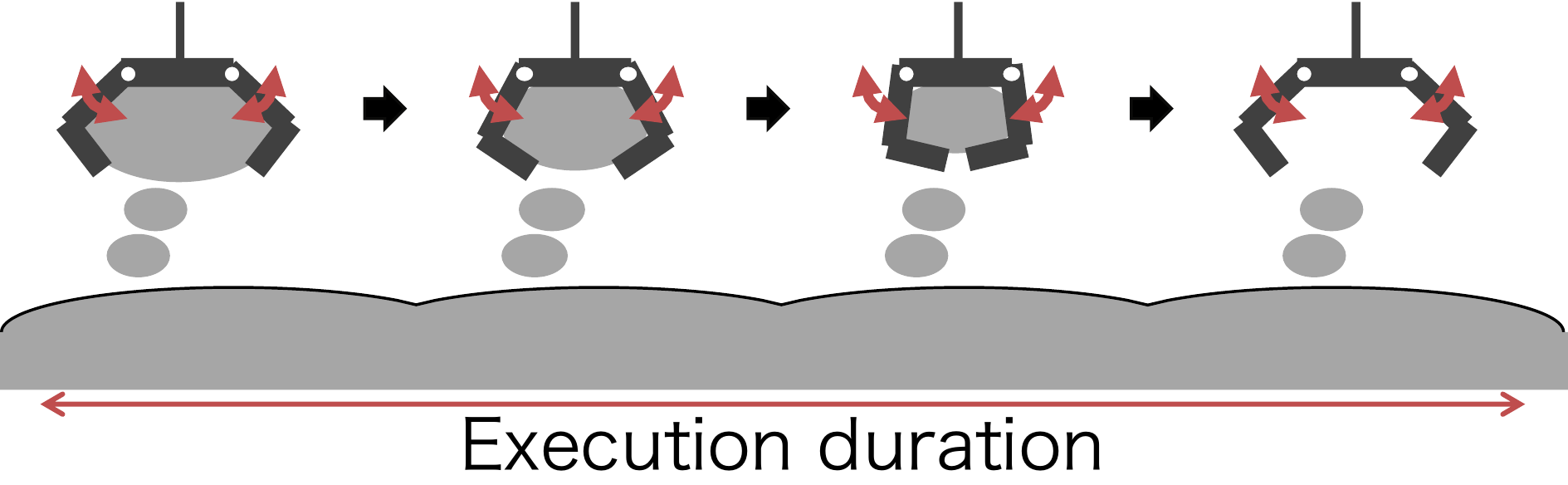} 
        \subcaption{Scattering}
        \label{fig_problem_scattering}
    \end{minipage}
    \caption{Garbage-grasping-scattering task}
    \label{fig_problem}
\end{figure}

\subsection{Garbage-grasping-scattering Task}
The garbage-grasping-scattering task aims to evenly and widely scatter a large amount of garbage in a short time (Fig. \ref{fig_problem}).
The grasping strategy is set as a motion that lowers the bucket onto the garbage's surface and closes the claws while raising the bucket.
The scattering strategy is set as a motion where some of the garbage in the bucket falls by opening and closing it.
To achieve the aim of the task, we need to switch between grasping and scattering, as well as their execution duration based on the garbage's characteristics

In this task, state $s_t$ is defined as the weight of the grasped garbage.
The action is defined as binary $a_t=\{0, 1\}$, which indicates either grasping or scattering strategies.
Execution duration $\tau_t$ indicates the number of execution steps for each strategy.
The reward function is defined:
\begin{align}
    r_t = 
    \begin{cases}
        0                   & (a_t=0)  \\
        r_a \times r_\tau   & (a_t=1) 
    \end{cases}.
\end{align}
$r_a$ is the action reward for the scattering performance, and $r_\tau$ is the reward related to a scattering's duration.
Each episode terminates when the crane finishes scattering the garbage. % grasped by the bucket. 

\begin{figure*}
    \centering
    \begin{minipage}[b]{0.8\linewidth}
        \centering
        \includegraphics[width=\hsize]{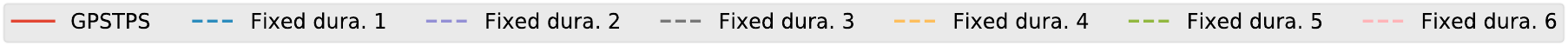}
    \end{minipage} \\
    \begin{minipage}[b]{0.24\linewidth}
        \centering
        \includegraphics[width=0.9\hsize]{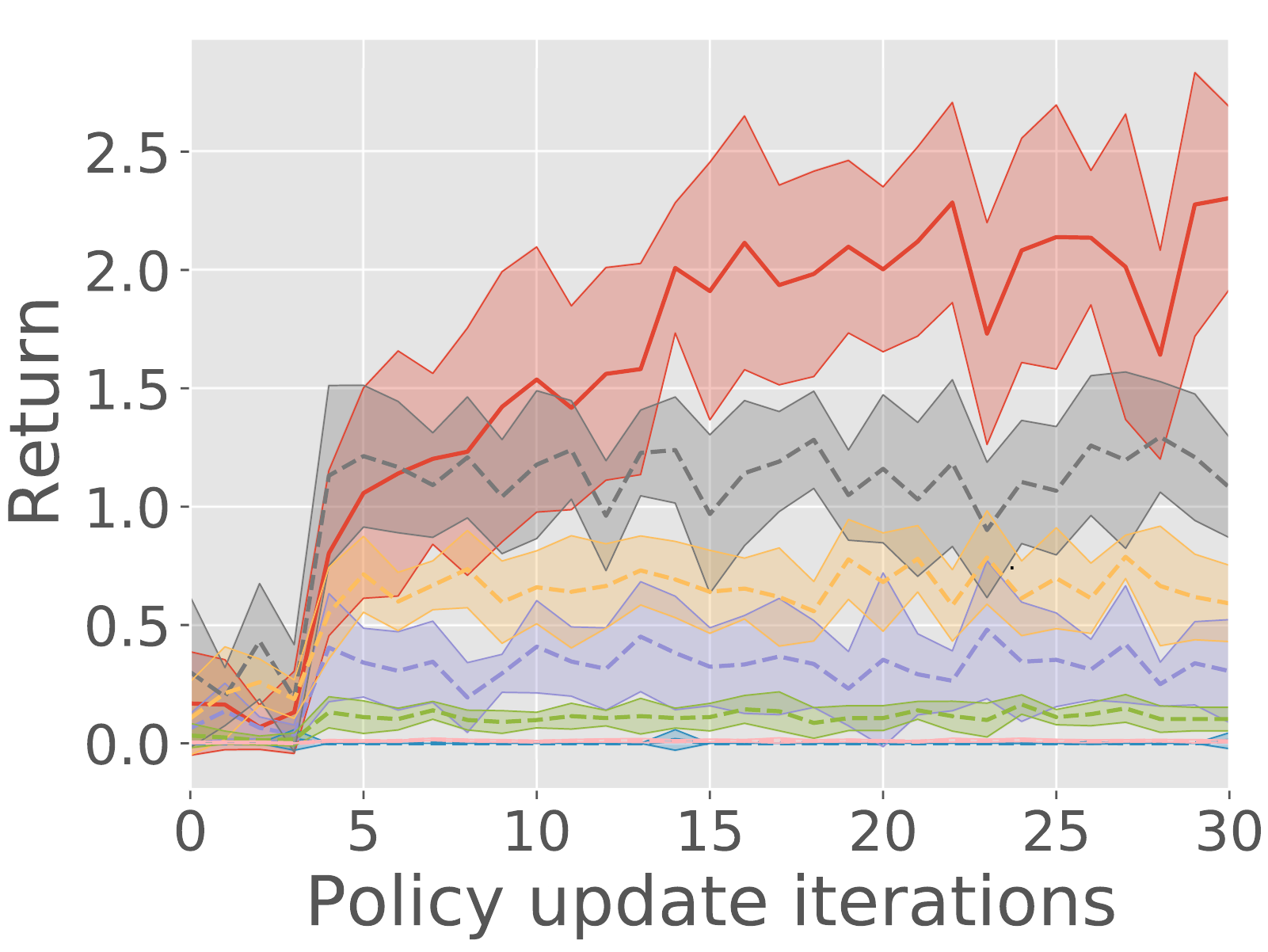}
        \subcaption{Learning curve}
        \label{fig_sim1_reward}
    \end{minipage} 
    \begin{minipage}[b]{0.24\linewidth}
        \centering
        \includegraphics[width=0.9\hsize]{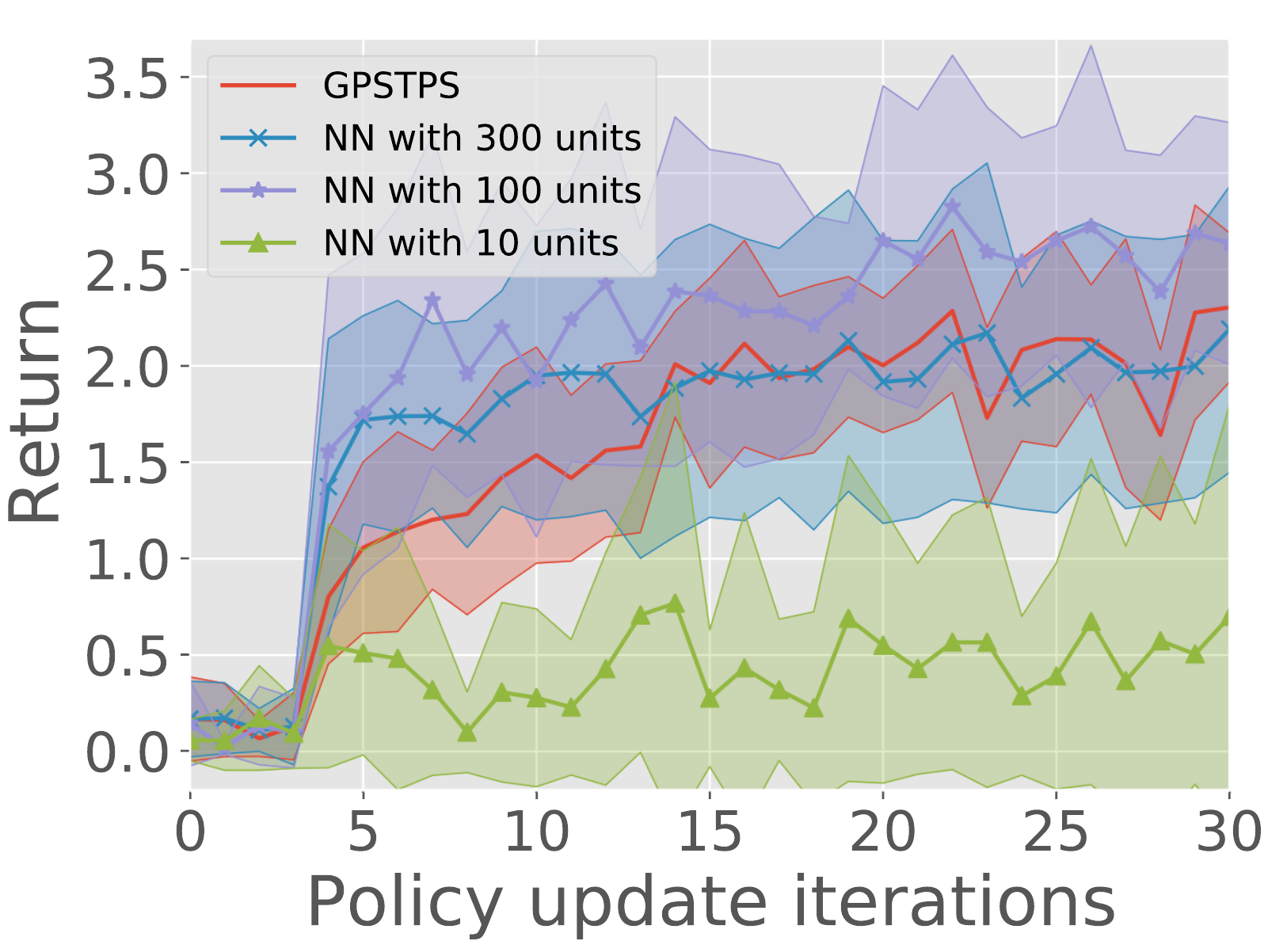}
        \subcaption{Learning curve}
        \label{fig_sim1_reward_NN}
    \end{minipage} 
    \begin{minipage}[b]{0.24\linewidth}
        \centering
        \includegraphics[width=0.9\hsize]{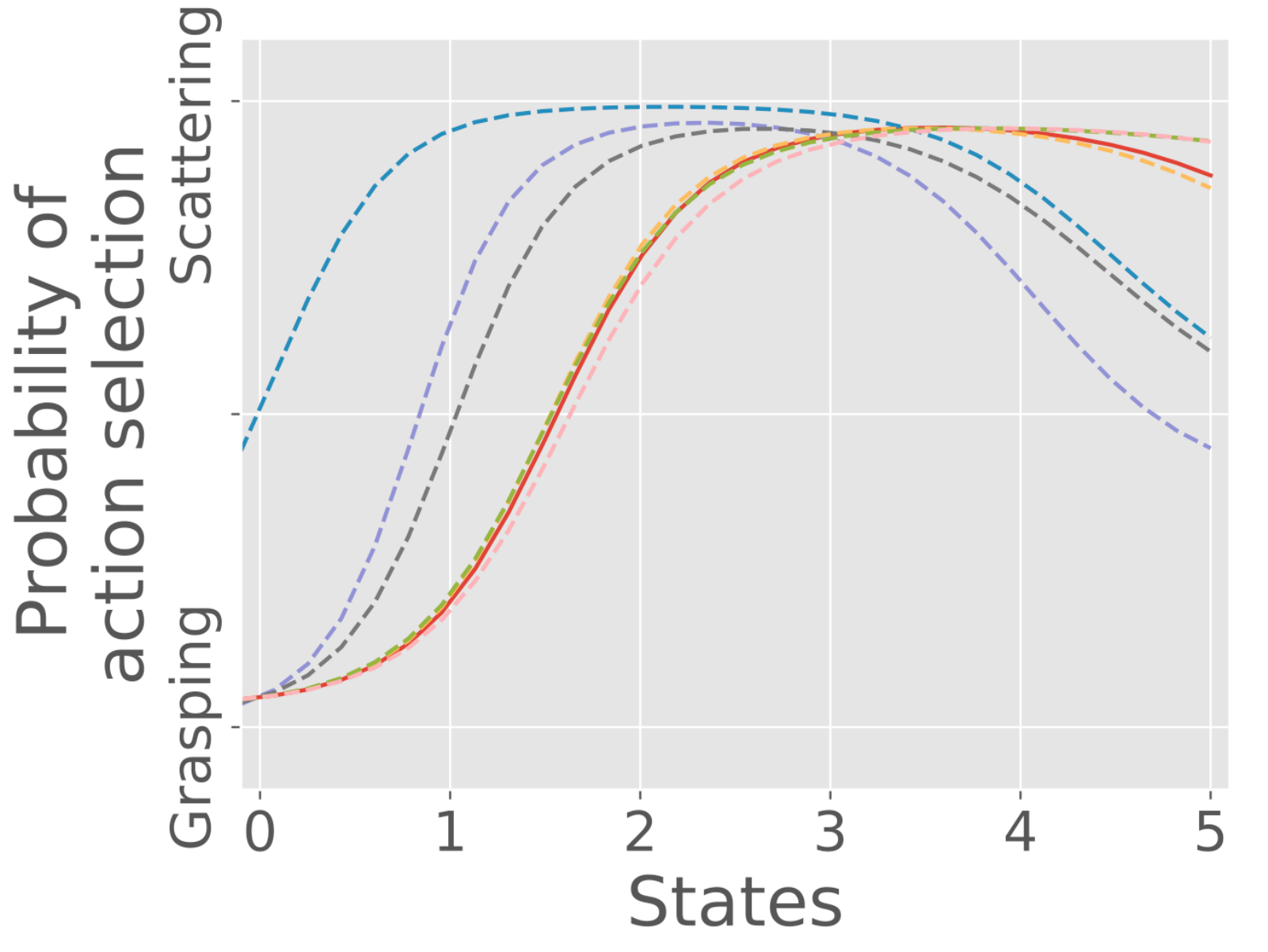} 
        \subcaption{Action policy}
        \label{fig_sim1_action_policy}
    \end{minipage}
    \begin{minipage}[b]{0.24\linewidth}
        \centering
        \includegraphics[width=0.9\hsize]{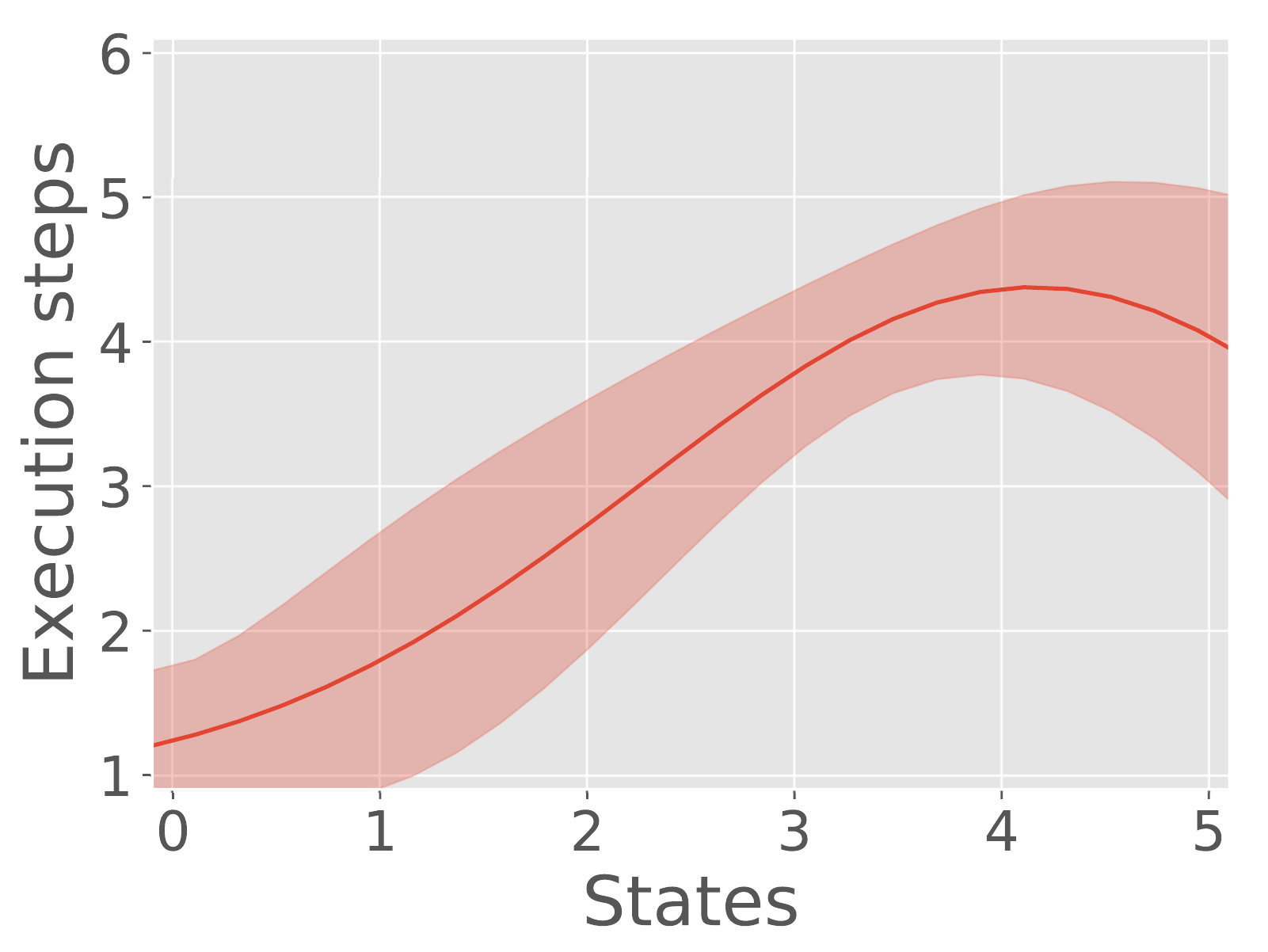} 
        \subcaption{Duration policy}
        \label{fig_sim1_trigger_policy}
    \end{minipage}
    \begin{minipage}[b]{0.24\linewidth}
        \centering
        \includegraphics[width=0.9\hsize]{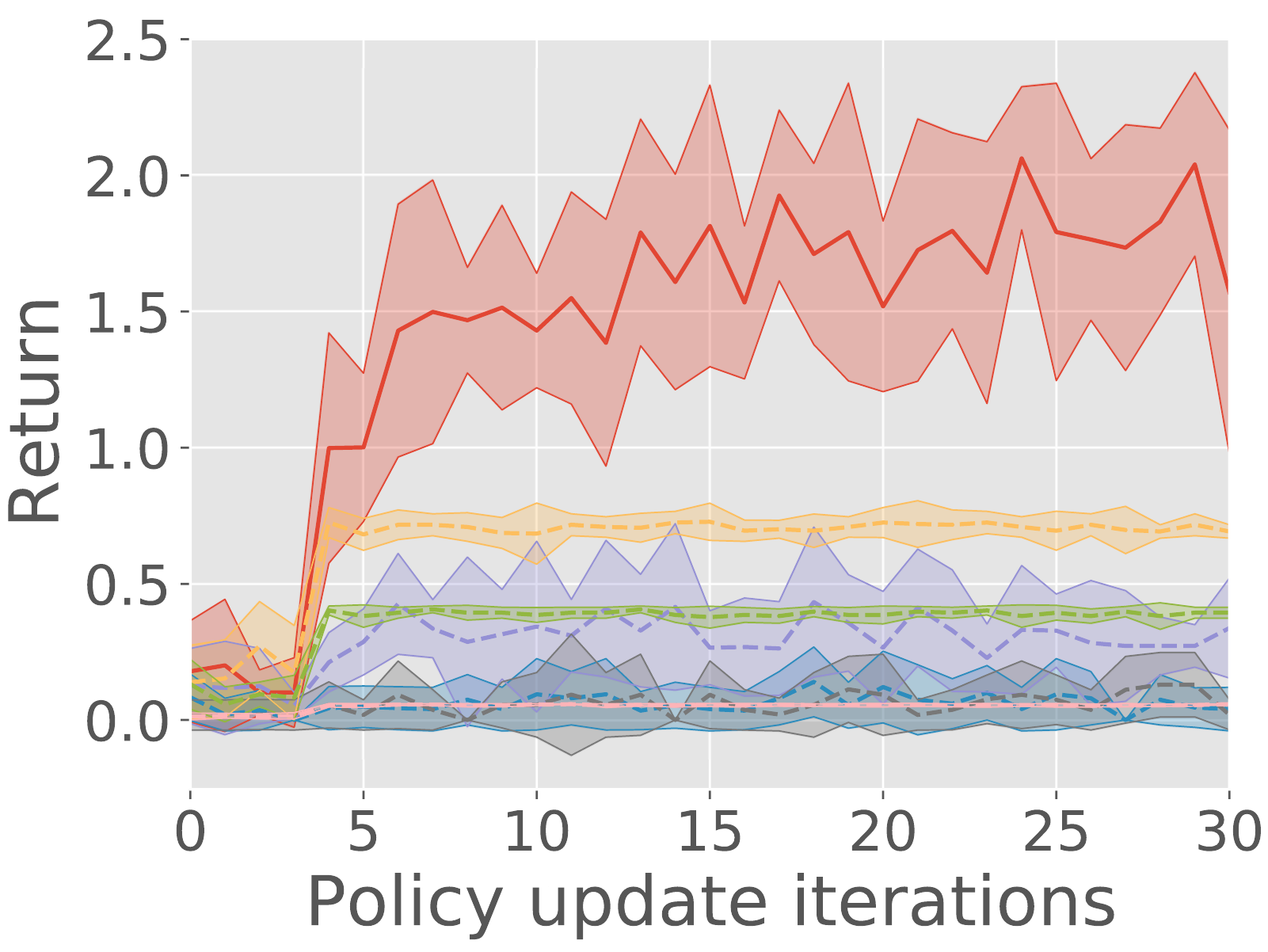}
        \subcaption{Learning curve}
        \label{fig_sim2_reward}
    \end{minipage} 
    \begin{minipage}[b]{0.24\linewidth}
        \centering
        \includegraphics[width=0.9\hsize]{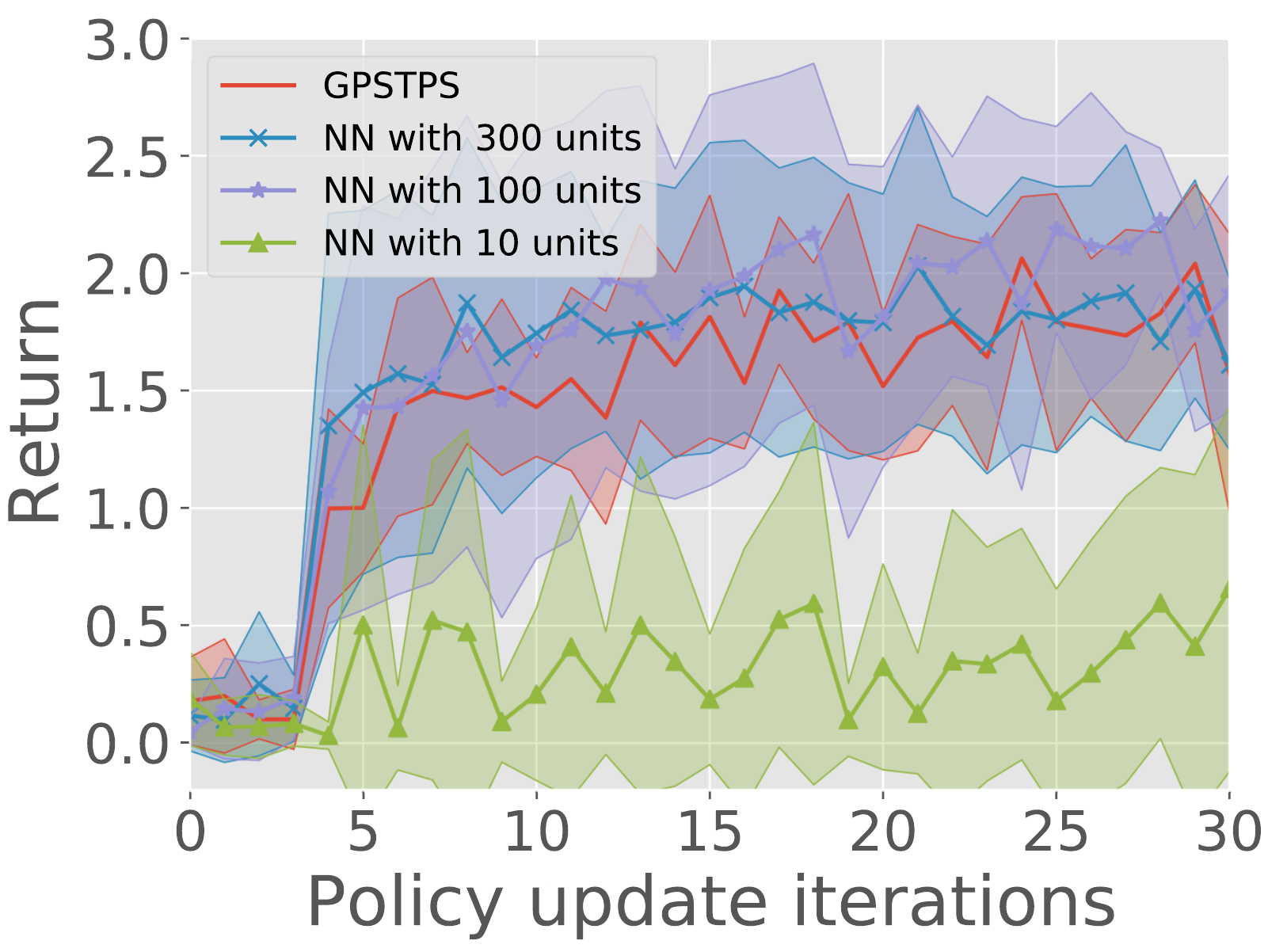}
        \subcaption{Learning curve}
        \label{fig_sim2_reward_NN}
    \end{minipage} 
    \begin{minipage}[b]{0.24\linewidth}
        \centering
        \includegraphics[width=0.9\hsize]{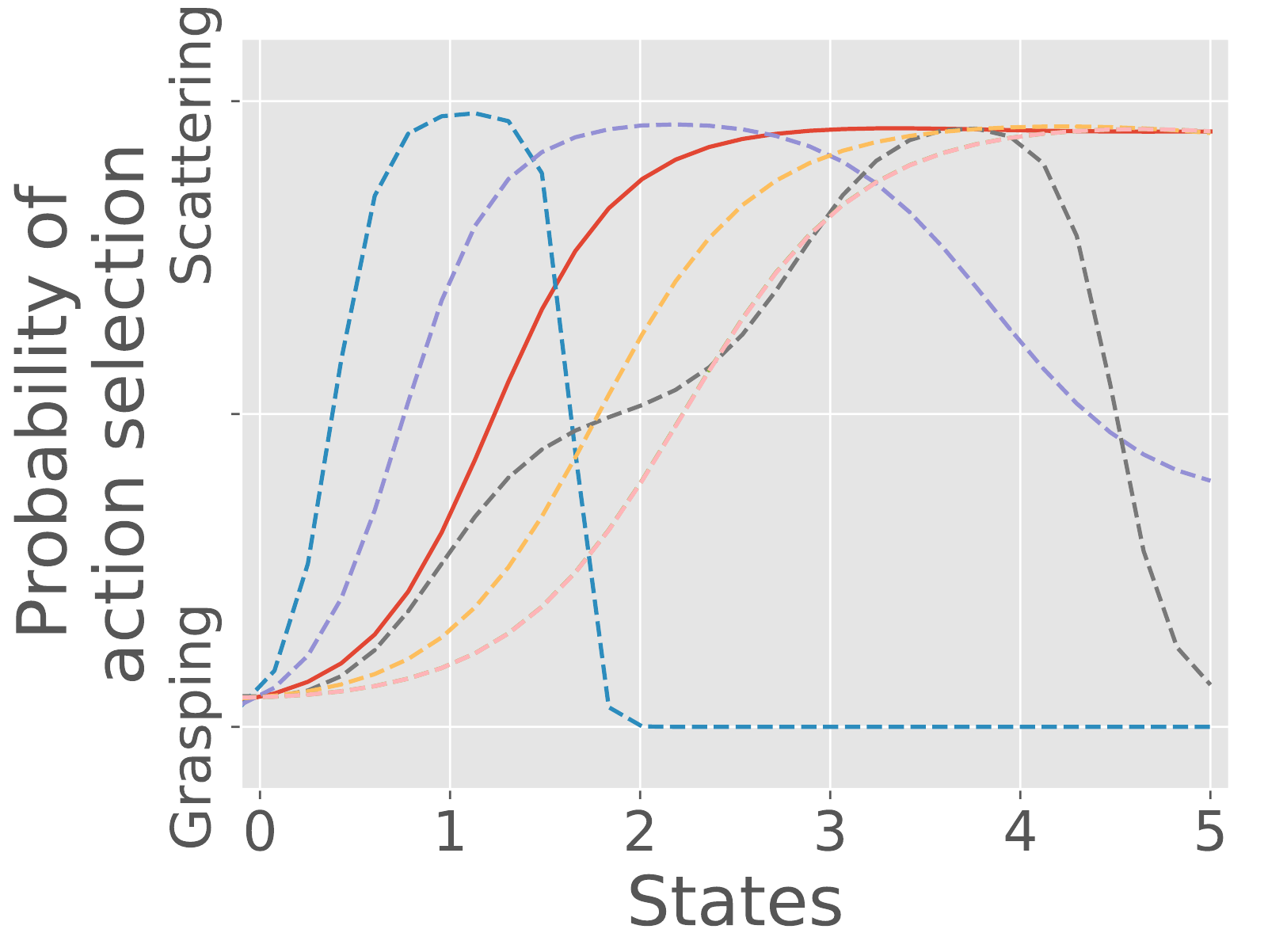} 
        \subcaption{Action policy}
        \label{fig_sim2_action_policy}
    \end{minipage}
    \begin{minipage}[b]{0.24\linewidth}
        \centering
        \includegraphics[width=0.9\hsize]{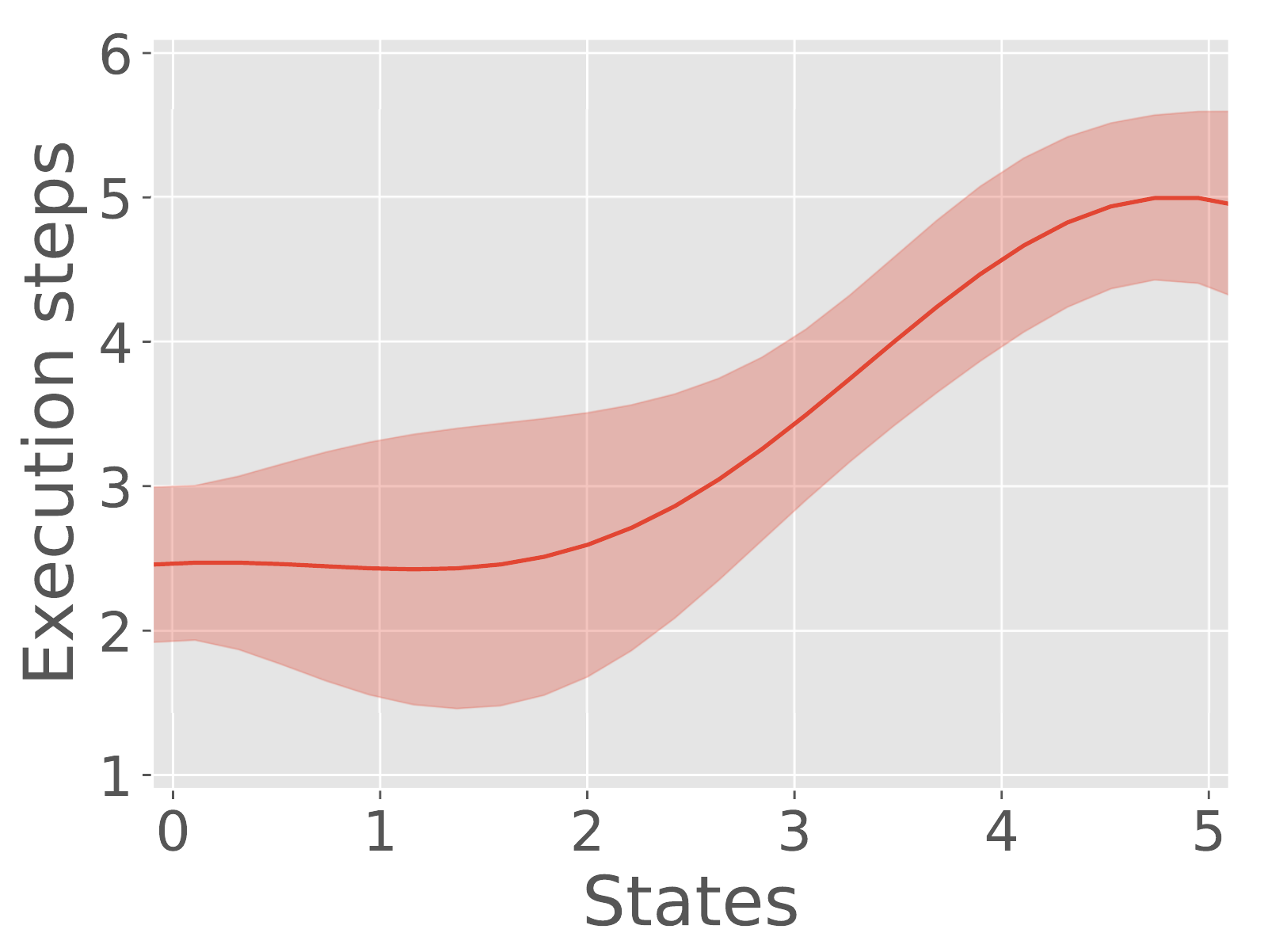} 
        \subcaption{Duration policy}
        \label{fig_sim2_trigger_policy}
    \end{minipage}
    \caption{Result of simulation experiments: (a)-(d) show results in setting 1. (e)-(h) show results in setting 2. (a) and (e) are the mean and standard deviation of the learning curve of ten experiments by GPSTPS and GPPS with a fixed duration. (b) and (f) mean and standard deviation of the learning curve of ten experiments by GPSTPS and STPS with NN. (c) and (g) learned action policies by GPSTPS. (d) and (h) learned duration policy by GPSTPS.}
    \label{fig_sim}
    \vspace{-3mm}
\end{figure*}

\subsection{Simulation}

\subsubsection{Experimental settings}
We simulated garbage grasping and scattering and verified the GPSTPS performance.
The amount of garbage grasped by the waste crane depends on its characteristics.
Thus, we prepared the two garbage characteristics shown in Table \ref{table_sim_garbage}.
Settings 1 and 2 indicate soft and hard garbage that requires different execution durations.
Hard garbage's grasping strategy requires a longer duration since it cannot be loosened by being grasped. 
We assumed that for one execution duration step, a grasping strategy takes ten seconds and that a scattering strategy takes five seconds.

We defined the action and time rewards:
\begin{align}
    r_a &= \mathrm{min}(s_t, \tau_t)-\alpha_\mathrm{sim}\|s_t-\tau_t\|, \\
    r_\tau &= \mathrm{exp}\{-\beta_\mathrm{sim}(u_\mathrm{act}-u_\textrm{min})^2\},
\end{align}
where $\alpha_\mathrm{sim}=1.5$, $\beta_\mathrm{sim}=0.004$, and $u_\mathrm{min}=30$ are the parameters of the reward function and $u_\mathrm{act}$ is the seconds required during garbage scattering.
The action reward is high when the execution durations are similar to a state and a large amount of garbage is scattered.
Time reward $r_\tau$ increases as the execution time is shortened.

The task begins without no grasped garbage in the bucket, and an episode terminates when the crane has scattered all of the grasped garbage.
The action policy is modeled by a binary sparse GP classification model with $m_f=0.5$.
Since the classification model directly regresses the probability of the action selection, we ignore the uncertainty of the predictive model.
The duration policy is sparse GP regression model with $m_g=0$.
The maximum execution duration is set to six steps.
For comparison, we employed GP policy search (GPPS) with a fixed duration from one to six steps.
Pseudo inputs of sparse GP in each method are set $M = 5$.
Also, we implemented a neural network (NN) based STPS, which employs NNs with three full-connect layers for both action and duration policies.
We set multiple numbers of units in a hidden layer in NN as 10, 100, and 300.
STPS with NN uses return weight likelihood derived in \cite{sasaki2021} as the objective function and learns each NN policy by Adam optimizer.

\begin{table}
    \centering
    \caption{Amount of garbage grasped with respect to grasping durations in simulation experiment: $\epsilon$ is sampled by $\mathcal{N}(0, 0.7)$.}
    \label{table_sim_garbage}
    \vspace{-1mm}
    \begin{tabular}{|c||cccc|}\hline
        Grasping durations & 1 & 2 & 3 & later \\ \hline\hline
        Setting 1 (soft)         & 3$+\epsilon$ & 3$+\epsilon$ & 3$+\epsilon$ & 3$+\epsilon$ \\
        Setting 2 (hard)         & 2$+\epsilon$ & 3$+\epsilon$ & 5$+\epsilon$ & 5$+\epsilon$ \\ \hline
    \end{tabular}
\end{table}

\subsubsection{Result}
Fig. \ref{fig_sim} shows the experimental result.
Fig. \ref{fig_sim1_reward}, \ref{fig_sim1_reward_NN}, \ref{fig_sim2_reward}, and \ref{fig_sim2_reward_NN} compares learning curves of GPSTPS, GPPS with a fixed duration, and GPSTPS with NN policy in each setting.
Fig. \ref{fig_sim1_action_policy} shows that the action policy appropriately selected the grasping and scattering strategies based on the state.
In Fig. \ref{fig_sim1_trigger_policy}, the duration policy selected a short duration (a one- or two-step grasping strategy) since the amount of grasped garbage was not changed by the execution duration.
The duration policy selected a similar execution time step as a state for the scattering strategy.
The action policy learned by GPSTPS appropriately selected the action based on the state (Fig. \ref{fig_sim2_action_policy}).
The learned duration policy selected a longer two- or three-step execution duration than the duration policy in setting 1 as its grasping strategy (Fig. \ref{fig_sim2_trigger_policy}).
For the scattering strategy, this duration policy resembles the duration policy learned in setting 1. 

In the learning curve of GPPS with a fixed duration in both settings shown in Figs. \ref{fig_sim1_reward} and \ref{fig_sim2_reward}, we found different duration needs for each setting.
Our GPSTPS outperformed the fixed duration method by learning a suitable duration for each setting.
In comparison with the NN policy model, the performance of STPS with the NN policy model largely depends on the number of units in hidden layers. If we could appropriately set it (100 units), it would result in high performance; if we set too many (300 units) or too few (10 units), the performance will be severely degraded. 
On the other hand, our GPSTPS achieved comparable performance to that by NN policy models with appropriate settings without explicitly setting the number of units in hidden layers due to non-parametric characteristics of GPs.  

In summary, these simulation results indicated that GPSTPS can learn appropriate action and execution duration policies depending on the garbage's characteristics. 

\subsection{Robot Experiment}

\begin{figure}
    \centering
    \begin{minipage}[b]{0.49\linewidth}
        \centering
        \includegraphics[width=0.7\hsize]{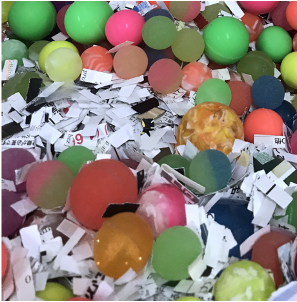} 
        \subcaption{Paper-based garbage}
        \label{fig_robot_garbage_paper}
    \end{minipage}
    \begin{minipage}[b]{0.49\linewidth}
        \centering
        \includegraphics[width=0.7\hsize]{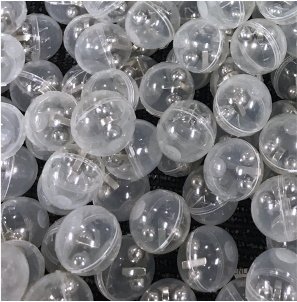} 
        \subcaption{Magnet-based garbage}
        \label{fig_robot_garbage_magnet}
    \end{minipage}
    % \caption{Environment of robot experiment: (a) overview, (b) bucket, (c), and (d) garbage with different characteristics.}
    \caption{The mock garbage with two different characteristics}
    \label{fig_robot_env}
\end{figure}

\begin{figure*}
    \centering
    \begin{minipage}[b]{0.8\linewidth}
        \centering
        \includegraphics[width=\hsize]{legend.pdf}
    \end{minipage} \\
    \begin{minipage}[b]{0.24\linewidth}
        \centering
        \includegraphics[width=0.9\hsize]{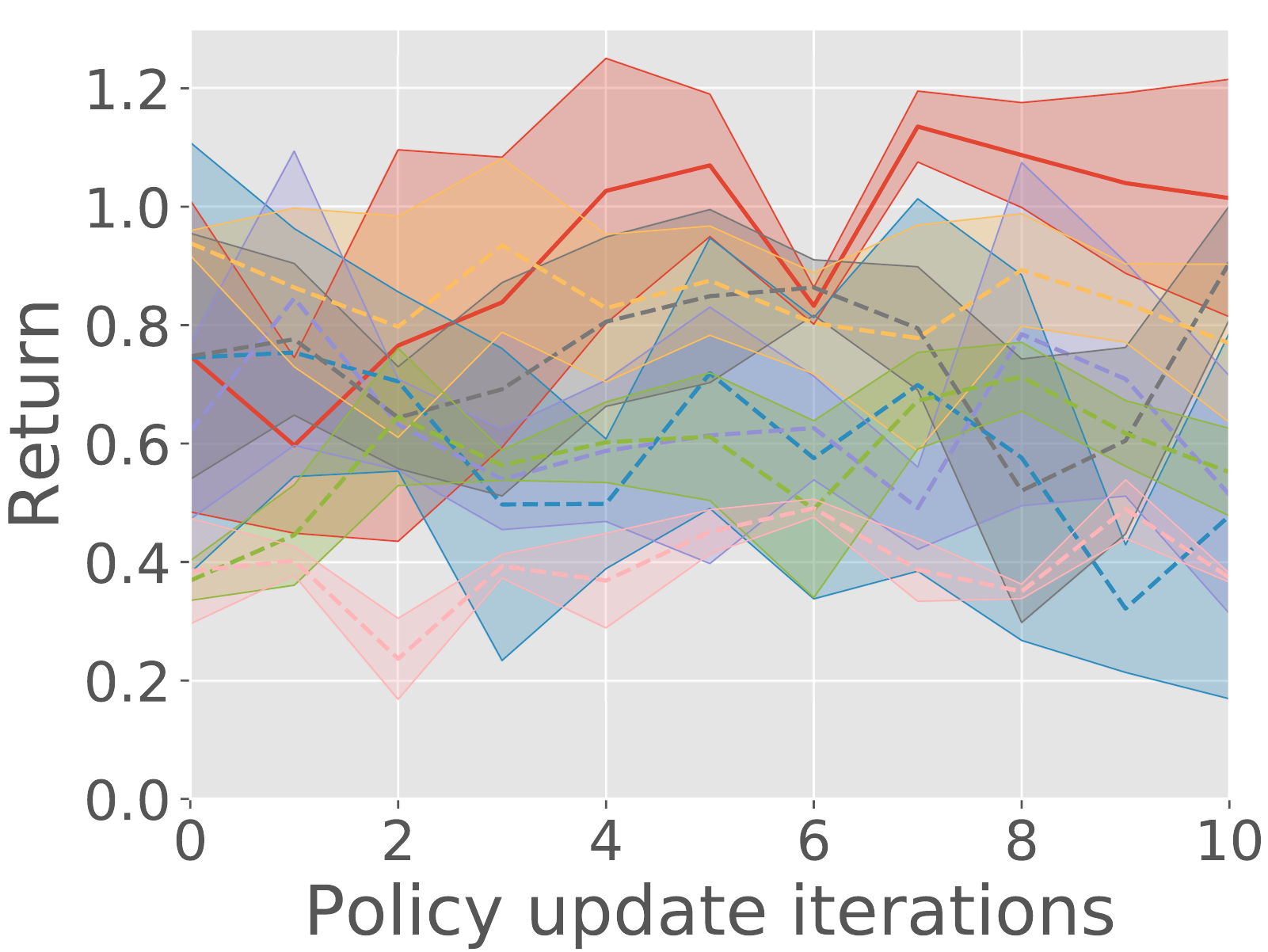}
        \subcaption{Learning curve}
        \label{fig_robot_paper_reward}
    \end{minipage}
    \begin{minipage}[b]{0.24\linewidth}
        \centering
        \includegraphics[width=0.9\hsize]{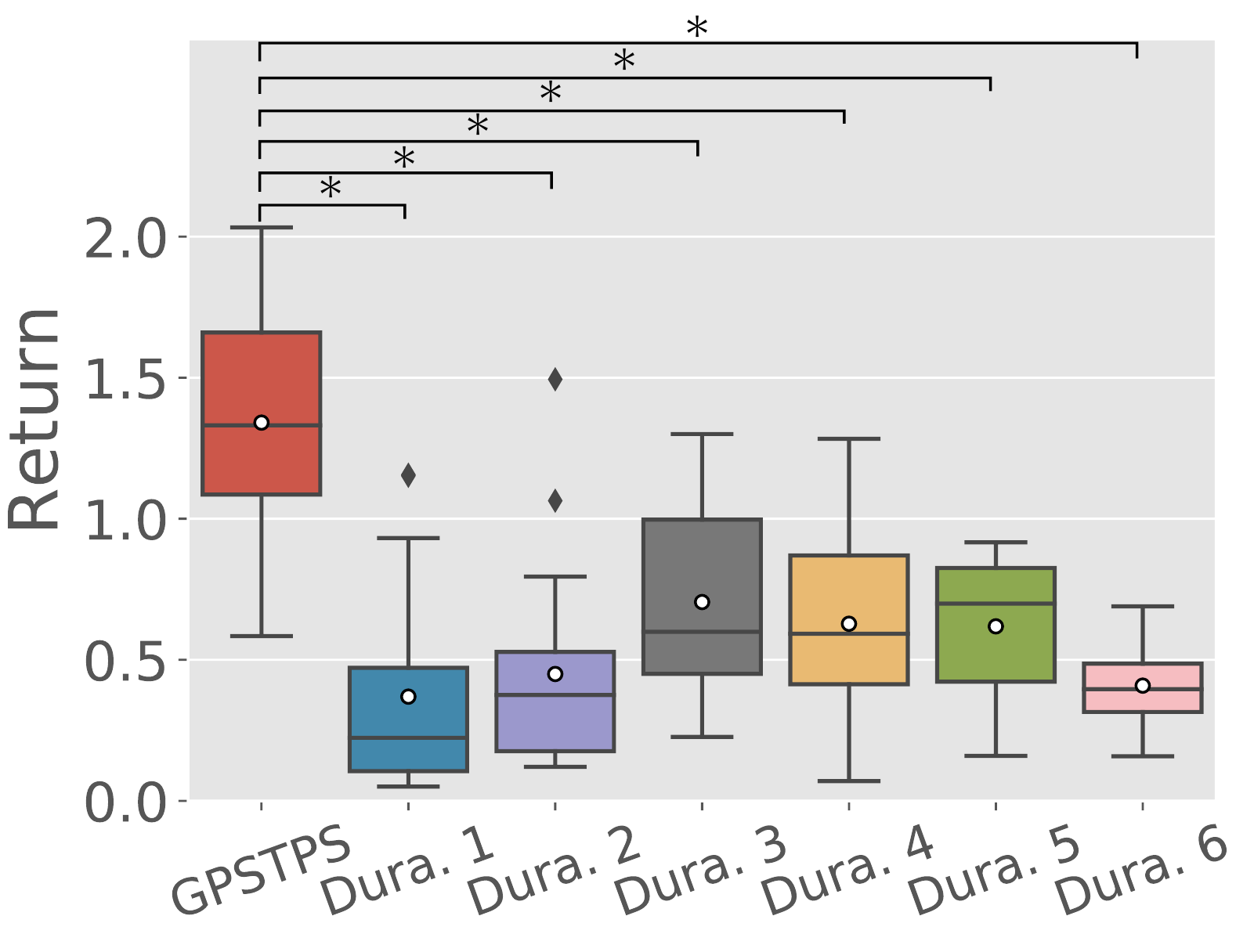}
        \subcaption{Return by learned policies}
        \label{fig_robot_paper_test}
    \end{minipage} 
    \begin{minipage}[b]{0.24\linewidth}
        \centering
        \includegraphics[width=0.9\hsize]{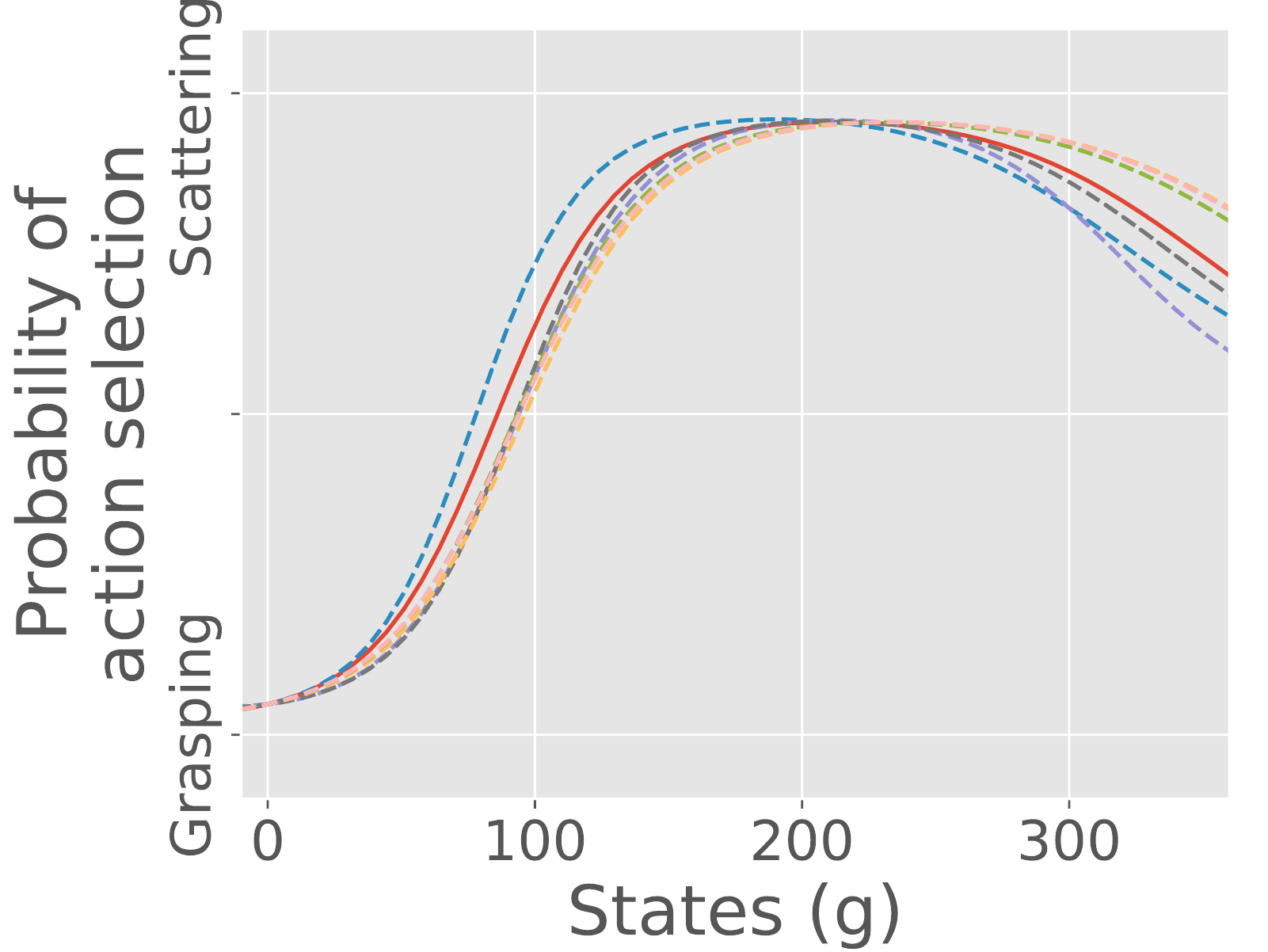} 
        \subcaption{Action policy}
        \label{fig_robot_paper_action_policy}
    \end{minipage} 
    \begin{minipage}[b]{0.24\linewidth}
        \centering
        \includegraphics[width=0.9\hsize]{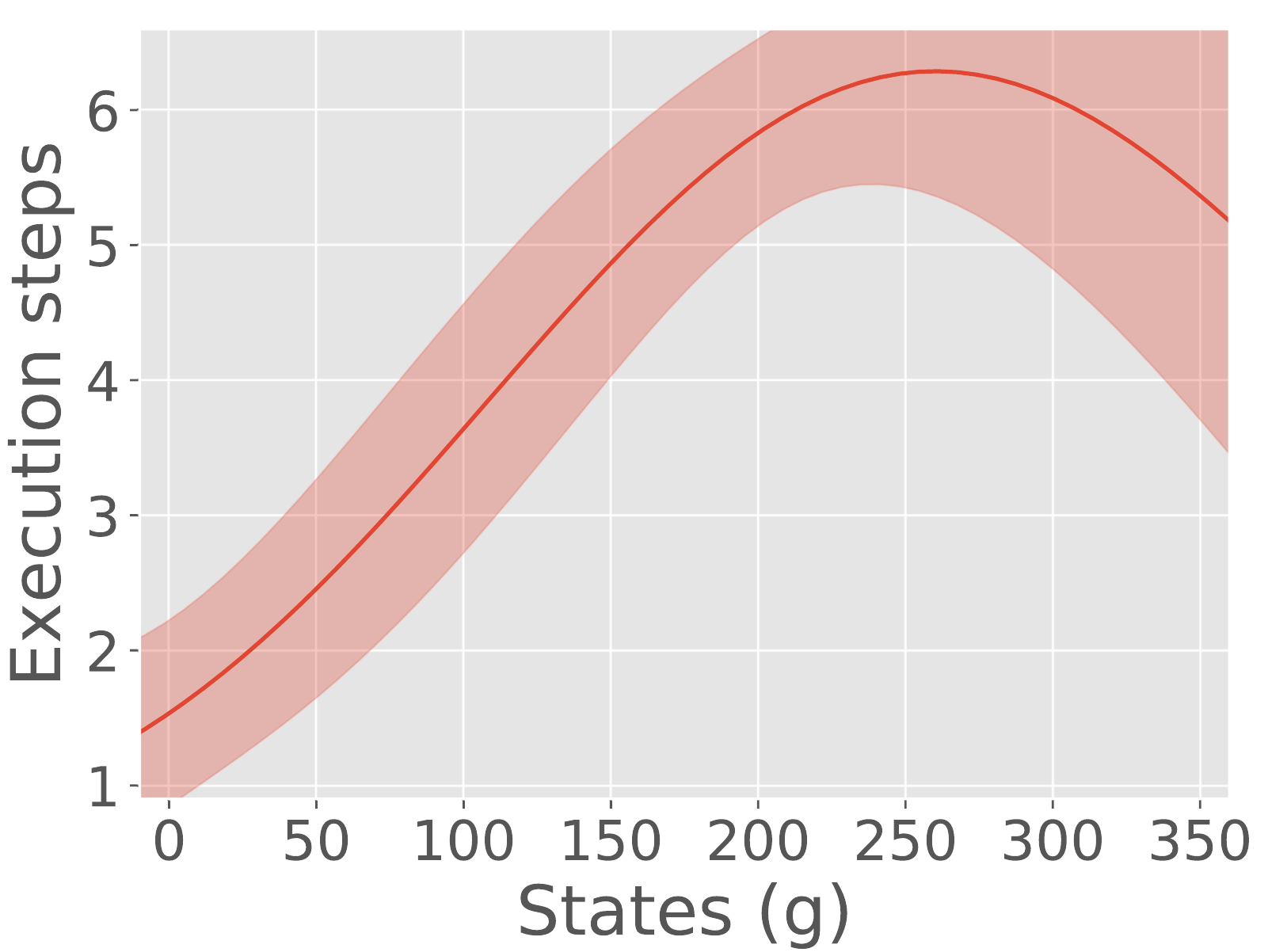} 
        \subcaption{Duration policy}
        \label{fig_robot_paper_trigger_policy}
    \end{minipage} \\
    \centering
    \begin{minipage}[b]{0.24\linewidth}
        \centering
        \includegraphics[width=0.9\hsize]{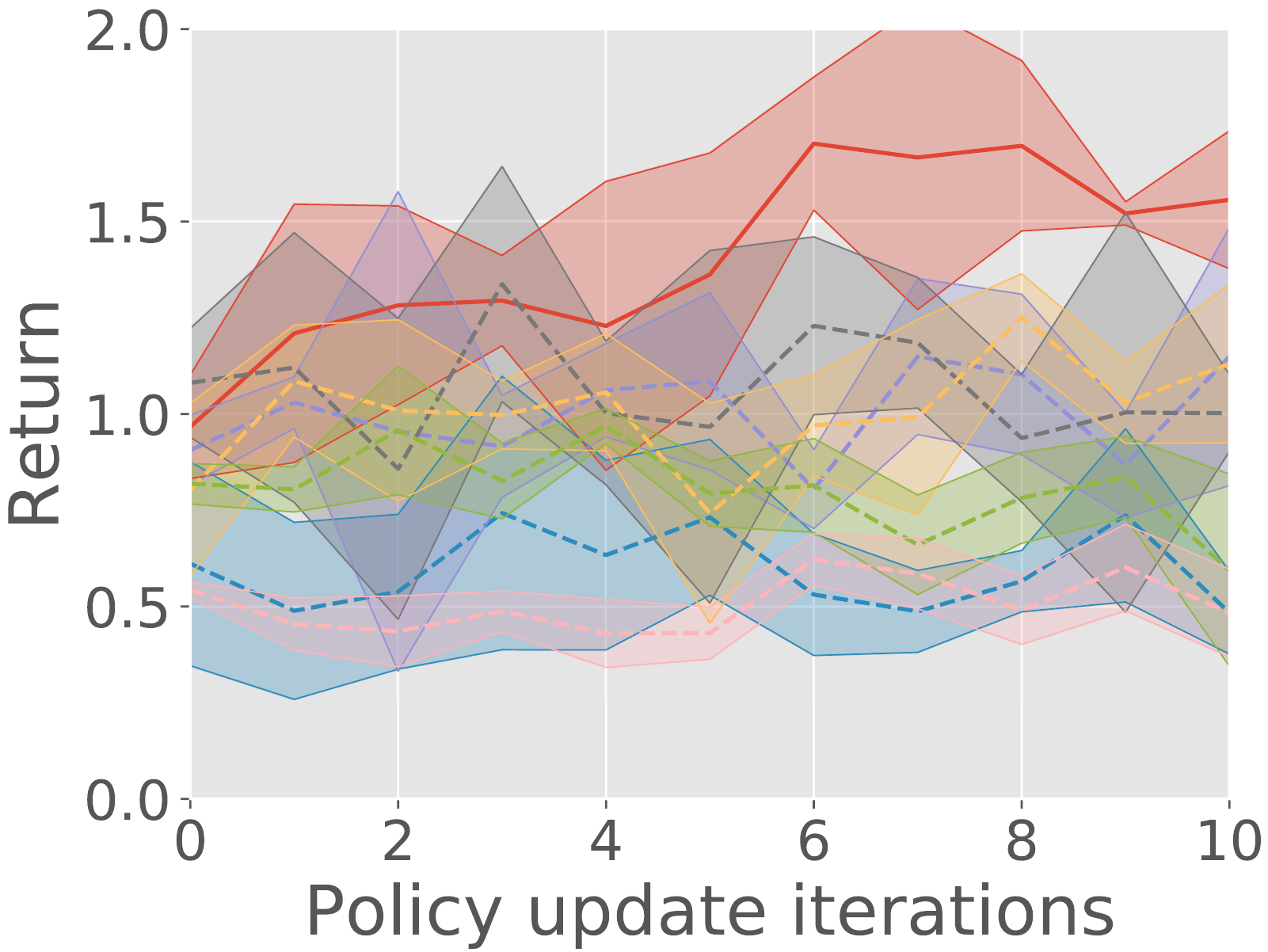}
        \subcaption{Learning curve}
        \label{fig_robot_magnet_reward}
    \end{minipage}
    \begin{minipage}[b]{0.24\linewidth}
        \centering
        \includegraphics[width=0.9\hsize]{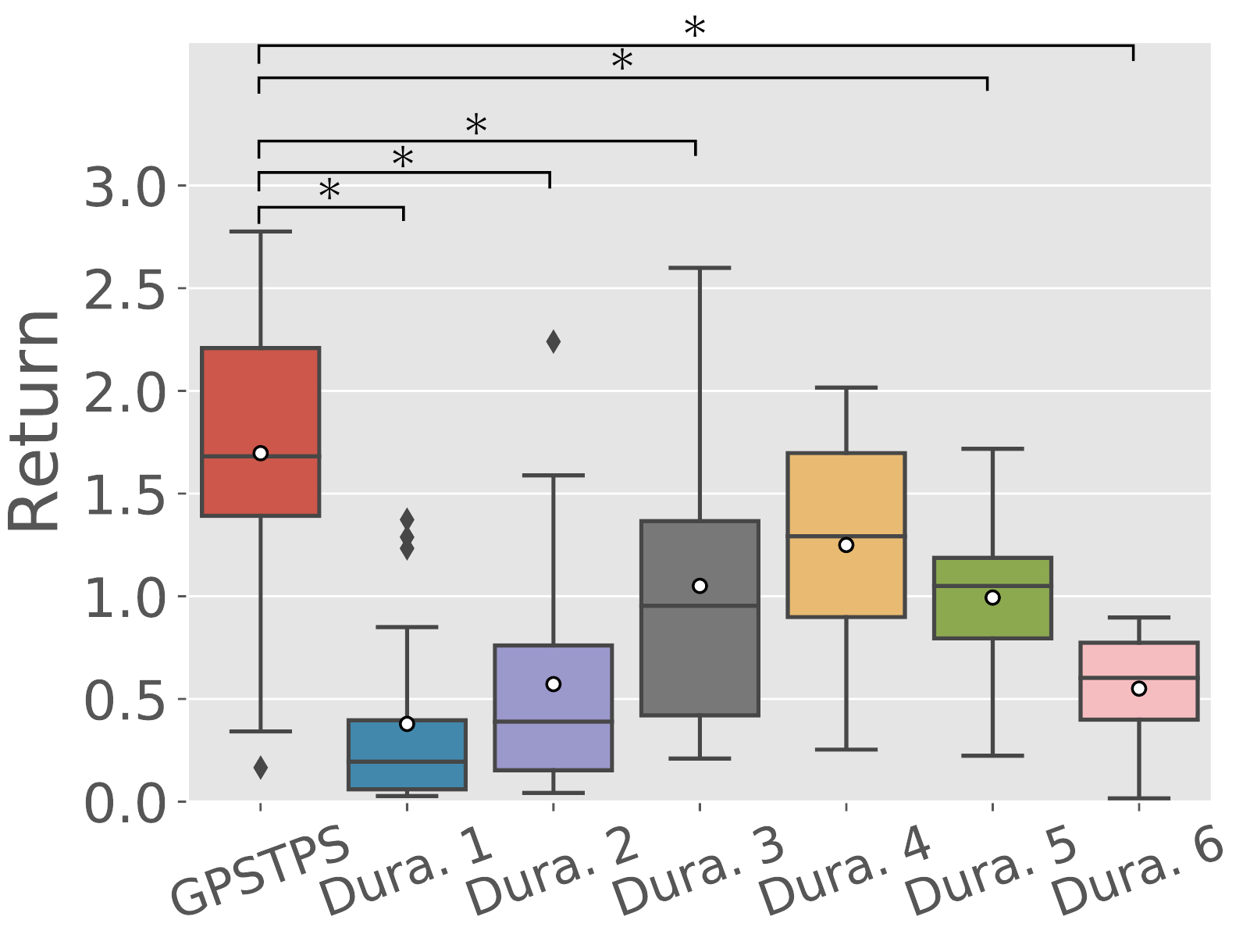}
        \subcaption{Return by learned policies}
        \label{fig_robot_magnet_test}
    \end{minipage}
    \begin{minipage}[b]{0.24\linewidth}
        \centering
        \includegraphics[width=0.9\hsize]{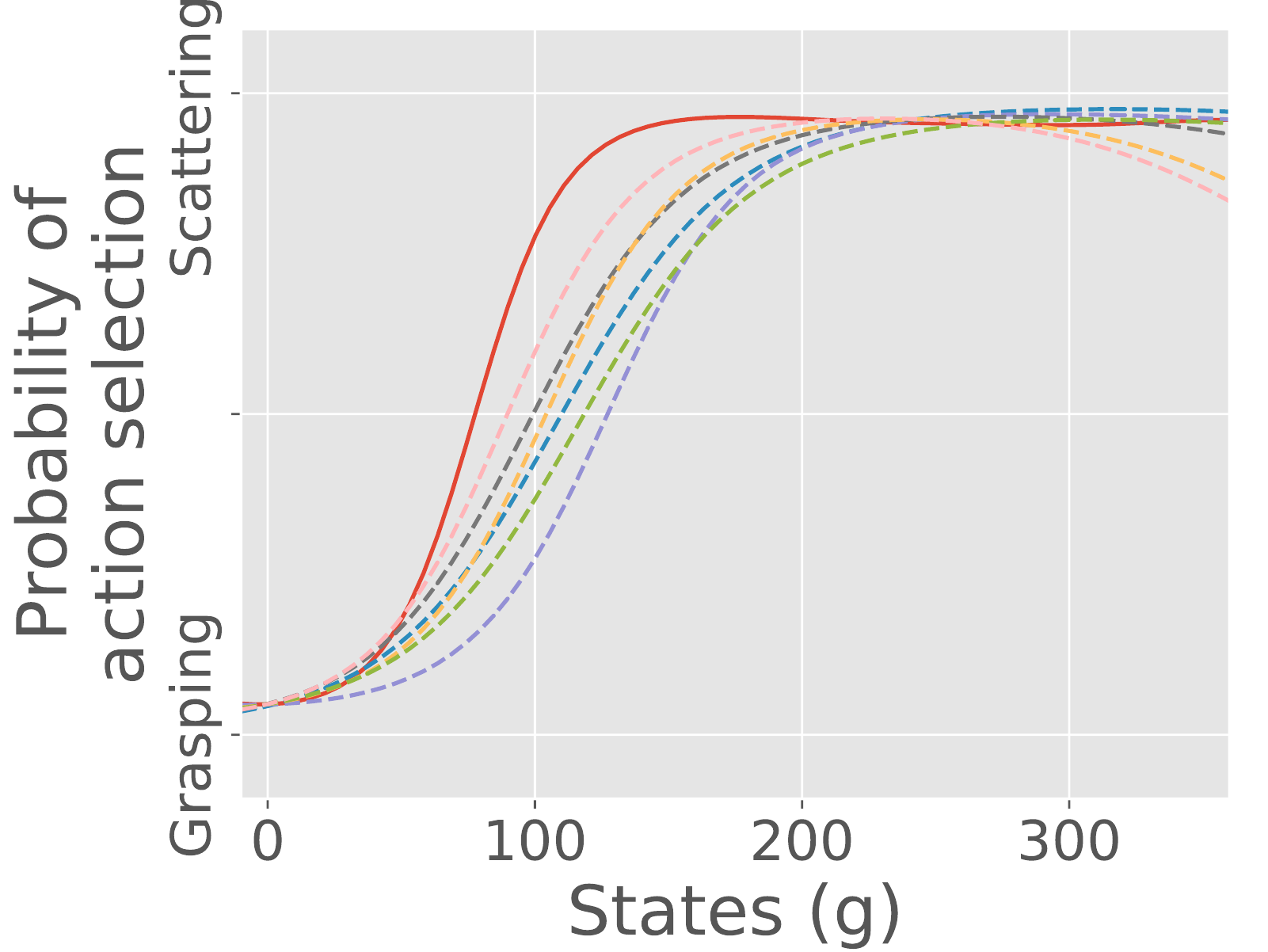} 
        \subcaption{Action policy}
        \label{fig_robot_magnet_action_policy}
    \end{minipage} 
    \begin{minipage}[b]{0.24\linewidth}
        \centering
        \includegraphics[width=0.9\hsize]{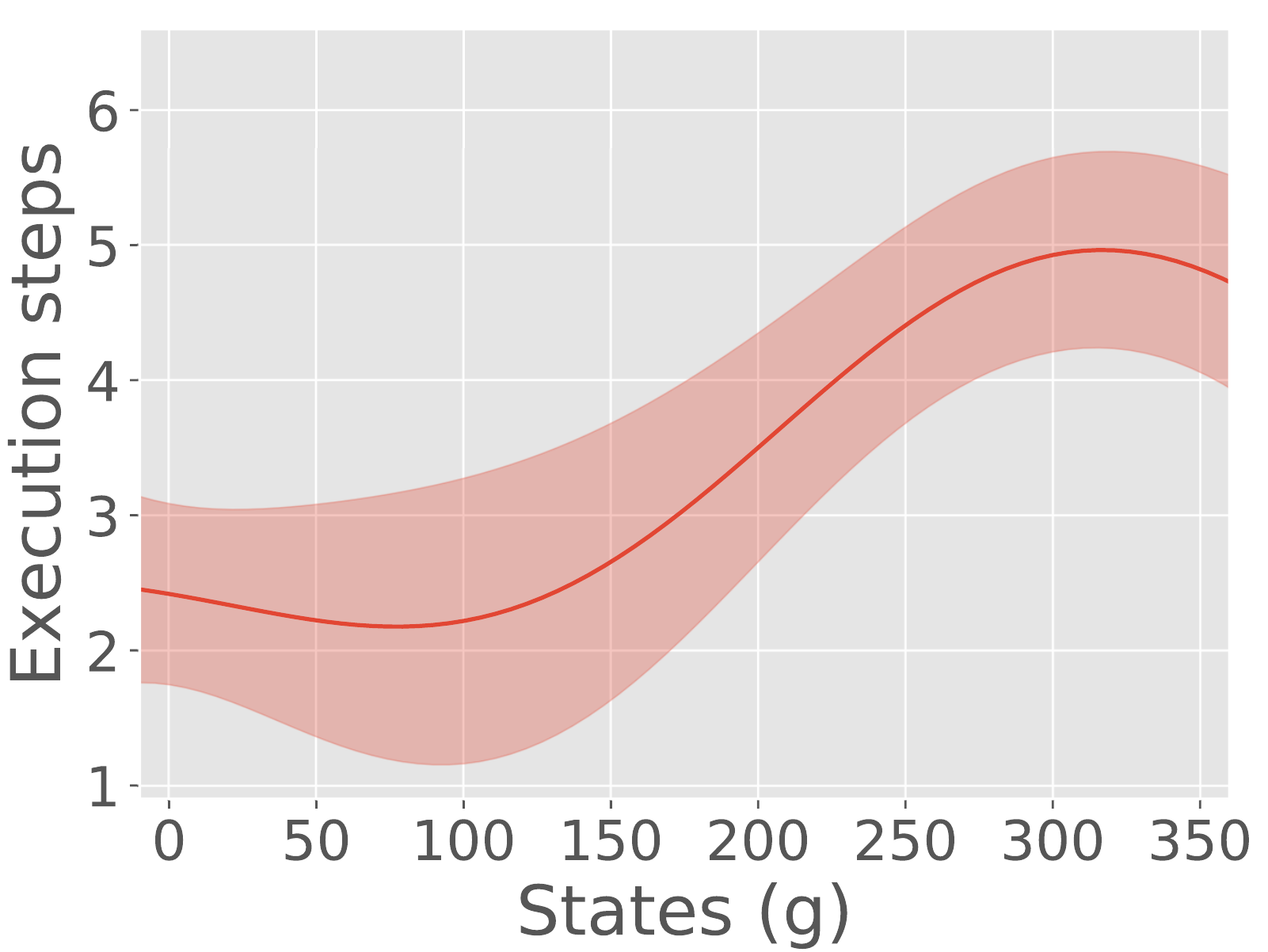} 
        \subcaption{Duration policy}
        \label{fig_robot_magnet_trigger_policy}
    \end{minipage} \\
    \caption{Result of robot experiment: (a)-(d) show results with paper-based garbage. (e)-(h) show results with magnet-based garbage. (a) and (e) mean and standard deviation of return of three experiments. (b) and (f) test performance of learned policies where * denotes $p<0.05$ on paired t-test. (c), (d), (g), and (h) action and duration policies learned by each method.}
    \label{fig_robot}
    \vspace{-2mm}
\end{figure*}

\begin{figure}
    \centering
    \begin{minipage}[b]{0.75\linewidth}
        \includegraphics[width=\hsize]{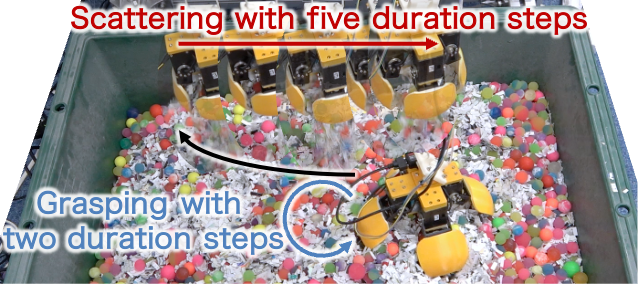}
        \subcaption{Paper-based garbage}
        \label{fig_robot_paper_trajecotry}
    \end{minipage} \\
    \begin{minipage}[b]{0.75\linewidth}
        \includegraphics[width=\hsize]{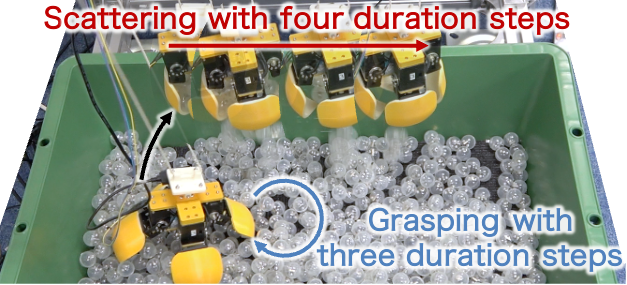}
        \subcaption{Magnet-based garbage}
        \label{fig_robot_magnet_trajectory}
    \end{minipage}
    \caption{Robot trajectory by policy learned by GPSTPS} 
\end{figure}

\subsubsection{Experimental settings}
We conducted an experiment with a robotic waste crane to verify GPSTPS's effectiveness. 
The detail of the robotic waste crane is described in \cite{sasaki2020}.
We used two types of garbage (Figs. \ref{fig_robot_garbage_paper} and \ref{fig_robot_garbage_magnet}) to verify the performance with different garbage characteristics.
Fig. \ref{fig_robot_garbage_paper} shows paper-based garbage that consists of shredded paper and rubber balls with 17 and 30 mm diameters.
Since this garbage is soft with a small particle size, it resembles dry and non-sticky garbage like plastic.
The bucket can grasp paper-based garbage with a short duration and the grasped garbage falls from a small gap between the claws.
The magnet-based garbage in Fig. \ref{fig_robot_garbage_magnet} is composed of 27-mm diameter capsules containing magnets and iron balls.
This garbage has a large particle size, and the magnets attract each other.
It resembles wet and easily aggregated garbage that is collected on rainy days.
The bucket needs a longer grasp duration because the garbage falls from the bucket during the aggregation.
The initial positions of the robotic crane in the grasping and scattering strategies are randomly selected in the garbage pit and automatically moved to the initial position.
The crane's moving distance in the scattering strategy is set to 30 cm.
% Each predetermined strategy is defined in Alg. \ref{alg_controller_grasping} and \ref{alg_controller_scattering}.
% The parameters in each algorithm are set to $t_\mathrm{lift}=0.8$ s, $w_\mathrm{fall}=7$ g, and $t_\mathrm{close}=2$ s.

In the robot experiment, action reward function $r_a$ and time reward function $r_\tau$ are defined:
\begin{align}
    r_a &= w_\mathrm{max}\mathrm{exp}\{-\gamma \mathrm{RMS}(\mathbf m-\mathbf m_\mathrm{I})\}, \\
    r_\tau &= \mathrm{exp}\{-\beta_\mathrm{robot}(u_\mathrm{act}-u_\mathrm{min})^2\},
\end{align}
where $\beta_\mathrm{robot}=2.5\times 10^{-4}$, $\gamma=7$, $u_\mathrm{min}=30$, and $u_\mathrm{act}$ is the execution time of an episode.
$u_\mathrm{min}$ means minimum execution time of grasping and scattering.
The purpose of this task is to scatter a lot of garbage in a shorter time evenly.
The action reward function evaluates scattering performance using RMS between sequence of grasped weight $\mathbf m$ and ideal weight $\mathbf m_\mathrm{I}$ that decreases weight linearly.
% The actual and ideal grasped weight sequences are shown in Fig \ref{fig_eval_func}.
The time reward function evaluates the shortness of scattering.

The policy is learned with the same experimental settings as in the simulation experiment.
The initial action policy selects the grasping strategy when the grasped weight is 0 g and the scattering strategy at other times.

\subsubsection{Result}
Fig. \ref{fig_robot} shows the experimental result of the robotic experiment.
Figs. \ref{fig_robot_paper_reward} and \ref{fig_robot_paper_test} show that GPSTPS outperformed GPPS with each fixed duration.
The paired t-test result shows that GPSTPS's return has a significant difference against all the compared methods. 
Figs. \ref{fig_robot_paper_action_policy} and \ref{fig_robot_paper_trigger_policy} indicate the learned policies.
Figs. \ref{fig_robot_paper_trajecotry} show the robotic waste crane trajectory by the policies learned by GPSTPS.
The learned action policy appropriately selected a control strategy.
The learned duration policy selected two and five steps of execution duration for grasping and scattering.
This result indicates that the duration policy captured the paper-based garbage's easy-to-grasp and gradually falling characteristics from a small gap between the claws.

Figs. \ref{fig_robot_magnet_reward} and \ref{fig_robot_magnet_test} show that GPSTPS outperformed GPPS with each fixed duration.
The paired t-test result shows that GPSTPS's return has a significant difference against all the compared methods except GPPS with a fixed four duration.
GPSTPS and GPPS with such a duration do not have a significant difference, although GPSTPS obtained higher returns than GPPS with the fixed four duration. 
Figs. \ref{fig_robot_magnet_action_policy} and \ref{fig_robot_magnet_trigger_policy} show the learned policies.
Fig. \ref{fig_robot_magnet_trajectory} and the robotic waste crane trajectory by the policy learned by GPSTPS.
The learned duration policy selected three duration steps for grasping the garbage and four for scattering it by capturing the characteristic of the magnet-based garbage of the difficult-to-grasp and falling together.
In the robotics experiment, running each episode took about two to three minutes.
One learning experiment took three hours.

Table \ref{table_robot_reward} shows the performance of initial and learned policies by GPSTPS in terms of RMS weight sequence and task execution time. We confirmed that learned policy by GPSTPS significantly improved the evenness of scattering and execution time by t-test. 

\begin{table}
    \centering
    \caption{Comparison of the performance of initial and learned policy by GPSTPS. Each policy is compared in terms of RMS weight sequence and the execution time of an episode. Values indicate mean and standard deviation of reward by 20 times of execution of the garbage-grasping-scattering task with paper- and magnet-based garbage. * and ** denote $p<0.05$ and $p<0.01$ on paired t-test, respectively.}
    \label{table_robot_reward}
    \vspace{-2mm}
    \begin{tabular}{|c|c||c|c|}\hline
                    & Policy  & Paper garbage              & Magnet garbage              \\ \hline\hline
        RMS of      & Initial & $1.17\pm 0.992$*           & $1.74\pm 1.33$**            \\ % \cline{2-4} 
        weight seq. & Learned & $\mathbf{0.516\pm 0.354}$* & $\mathbf{0.629\pm 0.694}$** \\ \hline % \hline
        Execution   & Initial & $64.0\pm 19.5$*            & $59.4\pm 14.2$*             \\ % \cline{2-4}
        time        & Learned & $\mathbf{50.9\pm 18.4}$*   & $\mathbf{50.8\pm 8.91}$*    \\ \hline
    \end{tabular}
\end{table}

In summary, we conducted experiments on a garbage-grasping-scattering task using two different kinds of mock garbage.
% We confirmed that 
GPSTPS properly learned the action and duration policies based on the garbage's characteristics.

\section{DISCUSSION}
A future task is indispensable that experimentally verifies the possibility of applications with an actual waste crane.
We must also verify the effectiveness of cranes/buckets in other weakly observable environments.
GPSTPS assumes a specific factorized action and duration policy model as $\pi_a(a_t\mid\mathbf s_t)\pi_\tau(\tau_t\mid\mathbf s_t)$; however, there are other possibilities in factorization, such as $\pi_a(a_t\mid\mathbf s_t)\pi_\tau(\tau_t\mid\mathbf s_t, a_t)$ so that the duration depends on the action for more complicated tasks. 
% Such an extension would be our future work. 

Combining a model-based policy learning approach \cite{deisenroth2015, hashimoto2020} with our method may further improve the sample efficiency.
Moreover, variational learning-based policy search may converge to local optima. 
To alleviate such an issue, we could alternatively use Markov chain Monte Carlo methods.

Since industrial machines are operated by humans, learning control strategies by imitation learning \cite{daniel2016} may be useful for a wide range of applications.
If the applied task has multimodal state transitions, its policy model can be extended by multimodality or robustness \cite{sasaki2021}.
In addition, deep kernel learning \cite{wilson2016} can be introduced if handling such high-dimensional data is needed.

\section{CONCLUSIONS}
This study aimed to automate machines in weakly observable environments in an industrial work place and proposed GPSTPS that can learn both action and duration policies to repeatedly perform the same action to overcome uncertainty.
Its effectiveness was experimentally verified with simulations and a robotic waste crane system.

\bibliographystyle{IEEEtran}
\bibliography{reference}
% \bibliography{RAL-2022}

\end{document}